\documentclass[letterpaper, 10 pt, conference]{ieeeconf}  

\IEEEoverridecommandlockouts                              

\usepackage[utf8]{inputenc} 
\usepackage[T1]{fontenc}    
\usepackage{hyperref}       
\usepackage{url}            
\usepackage{booktabs}       
\usepackage{amsfonts}       
\usepackage{nicefrac}       
\usepackage{microtype}      
\usepackage{xcolor}         
\usepackage{graphicx}
\usepackage{multirow} 
\usepackage{cite}
\usepackage{subcaption}

\usepackage{bbm}
\usepackage{pifont}

\newcommand*{\ShowNotes}{}
\definecolor{darkred}{rgb}{0.7,0.1,0.1}
\definecolor{darkgreen}{rgb}{0.1,0.7,0.1}
\definecolor{dblue}{rgb}{0.2,0.2,0.8}
\definecolor{maroon}{rgb}{0.76,.13,.28}
\definecolor{burntorange}{rgb}{0.81,.33,0}
\definecolor{cyan}{rgb}{0.0,0.7,0.94}
\definecolor{salmon}{rgb}{0.99,0.51,0.46}
\definecolor{green}{rgb}{0.03,0.91,0.43}
\definecolor{forestgreen}{rgb}{0.13,0.55,0.13}
\definecolor{grey}{rgb}{0.4,0.4,0.4}
\definecolor{purple}{rgb}{0.29,0,0.51}
\definecolor{crimson}{rgb}{0.86,0.08,0.24}

\ifdefined\ShowNotes
  \newcommand{\colornote}[3]{{\color{#1}\bf{#2: #3}\normalfont}}
\else
  \newcommand{\colornote}[3]{}
\fi

\title{Assistant Placement Aria:  A Benchmark for Egocentric Placement Assistance}

\author{Amir Belder$^{1}$, Gon\c{c}alo Dias Pais$^{2}$, Refael Vivanti$^{3}$, Omri Carmi$^{3}$, Daniel DeTone$^{3}$, \\ Oren Shrout$^{4}$, Ido Gattegno$^{3}$, and Ayellet Tal$^{4}$
\thanks{$^{1}$Amir Belder is with Reality Labs, Meta inc. and Technion, institute of technology, {\tt amirbelder@campus.technion.ac.il}}%
\thanks{$^{2}$Gon\c{c}alo Dias Pais was an intern at Reality Labs, Meta inc. and now with Sensei, Lisbon, Portugal, {\tt gpais@sensei.tech}}%
\thanks{$^{3}$ These authors are with Reality Labs, Meta inc.}
\thanks{$^{4}$ These authors are with Technion, institute of technology, Haifa, Israel}
}

\begin{document}

\maketitle
\thispagestyle{empty}
\pagestyle{empty}

\begin{abstract}
Human assistance in robotics spans around several tasks such as navigation, object manipulation, and placement, where a key challenge is selecting target destinations that align with human intentions or preferences.
We focus on this challenge in the context of Virtual Placement (VP), the task of identifying all plausible target locations given scene context and human-centric constraints.
This differs from traditional placement tasks that typically focus on a single, predefined target location.
The VP problem is complex, as it requires both global and local reasoning about the scene's geometry, semantics, and plausibility.
To address this gap, we introduce {\bf Assistant Placement Aria}, the first benchmark to explore diverse aspects of VP, including global, local, and human-centric constraints.
It contains both synthetic and real indoor scenes annotated for three tasks: (i)~2D Panel Placement, (ii)~Sitting Suggestion, and (iii)~TV Placement.
Each scene includes 2D images, a 3D point cloud, and a textual description of the objects within the scene.
By contributing this benchmark, we aim to encourage further research in this underexplored and challenging field that is critically dependent on relevant data.
We also evaluate several foundation models for object detection and segmentation on our benchmark.

\end{abstract}

\section{Introduction}
\label{sec:intro}

Human assistance is a central theme in a variety of robotics tasks in both industrial and household settings. 
Such tasks include object placement and manipulation, as well as navigation to pre-defined target destinations for assistive purposes~\cite{Sheng_2024_NYC_Indoor_VPR,9811600,10801373,9560853}. 
A key challenge in these scenarios is determining which target destinations best reflect human intentions or preferences, a problem that is non-trivial and remains relatively underexplored. 
In this work, we address this challenge in the context of assistive object placement.

In the robotics field, assistive navigation focuses on reaching a specific destination, while object placement involves transferring an object to a target location and ensuring it is positioned with the correct pose.
In contrast, object placement in computer vision typically refers to inserting a single foreground object into a background image at a suitable position and scale. 
This task introduces several challenges that can be grouped into two categories: (1) determining the appropriate size and a {\em single} placement for the object~\cite{zhou2022learning,meng2023interactiveobject,parihar2024text2place}, and (2) rendering the object realistically within the target image~\cite{Zhu_2023_CVPR,lee2018contextawaresynthesis}. 
In this work, we adopt the computer vision perspective, but frame it as suggestions or recommendations that can guide either a human or a robot.
We refer to this problem as Virtual Placement (VP). Unlike traditional placement tasks, VP focuses on a more semantic and human-centric question:
{\em “Given the context of a scene and its environment, where would a person place the object?”}
The objective is to identify {\em all plausible} placement locations that align with human preferences and contextual cues.

The VP problem requires considering human preferences while accounting for global and local physical constraints.
For example, when placing a TV screen, it is important to identify a comfortable viewing point (e.g., a sofa or a bed).
Global information is necessary to ensure that no objects obstruct the line of sight from the viewing point, while local information is required to identify vacant areas on the wall.

Some question-answering datasets contain a small number of placement-related questions~\cite{chen2020scanrefer,Peng_2023_CVPR,Ding_2023_CVPR,Azuma_2022_CVPR}, but we believe these are insufficient to capture the complexity of the placement problem.
In~\cite{ramrakhya2024seeingtheunseen}, a semantic (virtual) placement dataset for small man-made objects (e.g., table lamps and books) was introduced.
Although this dataset achieves impressive results, its annotations do not explicitly account for human preferences, as they rely solely on the locations where objects were observed.
For instance, if a book was not seen on a desk but only on bookshelves, a desk would never be considered a valid placement target.
Moreover, the dataset is limited to man-made objects, whereas assistive sitting introduces additional physical and human-centric constraints, such as comfort and accessibility.
For example, a person might avoid sitting on a crowded couch, while a book can be placed among other books without disrupting the scene’s natural appearance.
Due to these challenges, datasets covering diverse VP tasks remain scarce. 
This highlights the need for a new benchmark that supports broader exploration of VP constraints and enables further progress in the field.

\begin{figure*}[htbp]
  \centering
  \begin{subfigure}[t]{0.475\textwidth}
    \centering
    \includegraphics[width=\linewidth]{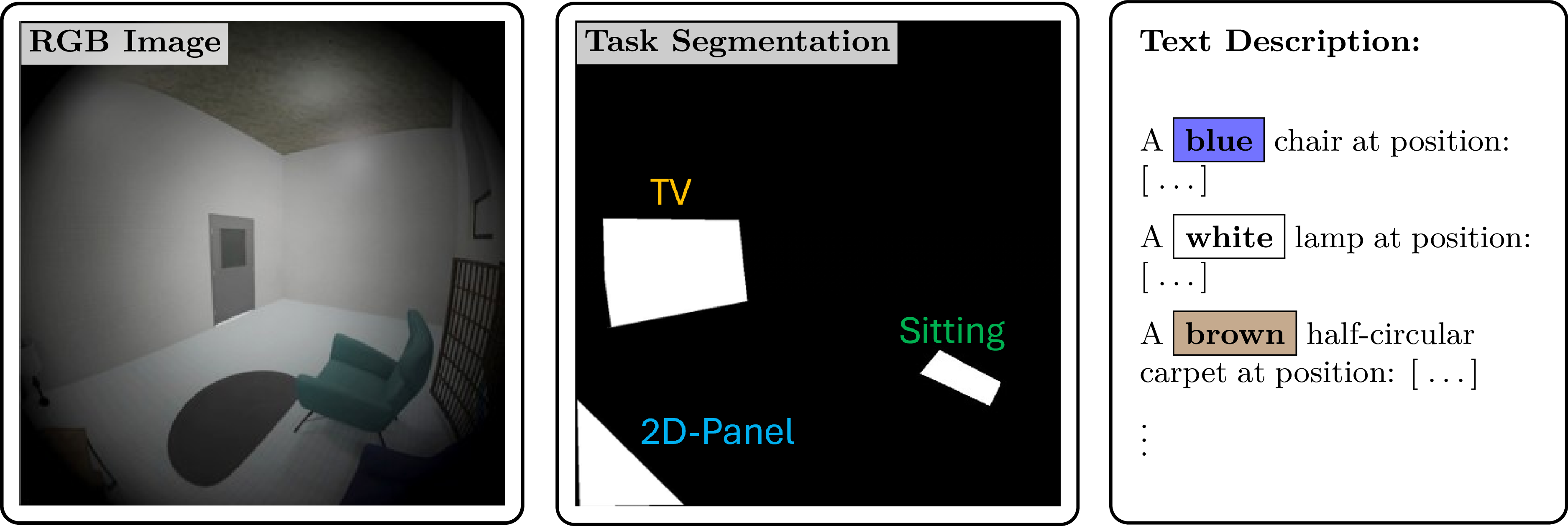}
    \caption{Synthetic}
    \label{fig:synthetic}
  \end{subfigure}
  \hfill
  \begin{subfigure}[t]{0.475\textwidth}
    \centering
    \includegraphics[width=\linewidth]{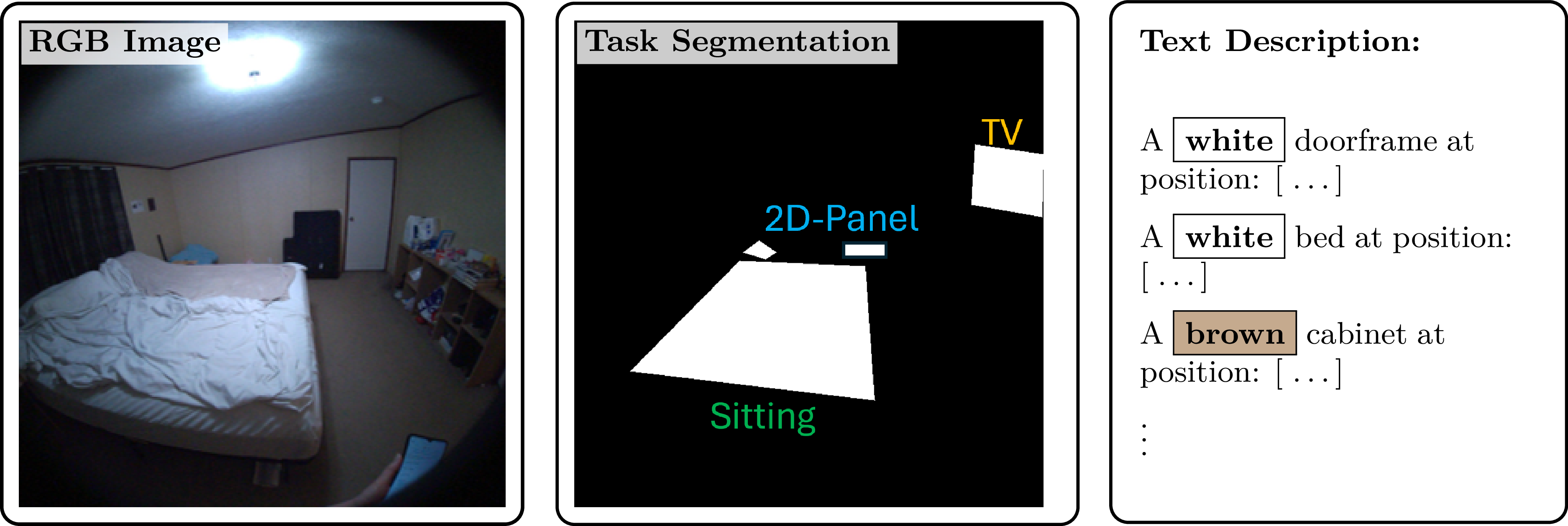}
    \caption{Real}
    \label{fig:real}
  \end{subfigure}
\caption{
{\bf Virtual placement.} 
Examples from our benchmark combining all three placement tasks in a single visualization: (a) a synthetic scene and (b) a real one. The tasks include (1) 2D Panel Placement, (2) Sitting Suggestion, and (3) TV Placement. Both are annotated with binary placement maps (white = placeable, black = non-placeable), and a VLM generates per-frame text descriptions including object characteristics and positions.
} 
\label{fig:teaser}
\end{figure*}

Thus, we introduce {\bf Assistant Placement Aria}, a novel VP benchmark annotated on the Aria Synthetic Environments (ASE)\cite{engel2023project} and Aria Everyday Objects (AEO)\cite{straub2024efm3dbenchmarkmeasuringprogress} datasets, covering three semantically relevant tasks:
(1) {\bf 2D Panel Placement} (e.g., a tablet);
(2) {\bf Sitting Suggestion}; and
(3) {\bf TV Placement}.
These tasks were chosen for their distinct characteristics, which together capture diverse aspects of VP. 
The first task, 2D Panel Placement, focuses on placing small man-made objects on different pieces of furniture. Since most such objects (e.g., books, table lamps, laptops, cups) typically require identifying vacant surfaces, we abstract them as a generic 2D horizontal panel with varying sizes to represent multiple object types.
The second task, Sitting Suggestion, requires considering comfort and accessibility, such as ensuring adequate legroom and avoiding overcrowded areas.
The third task, TV Placement, demands both global and local reasoning, as well as human-centric constraints, like, selecting a suitable height, locating available wall space, and ensuring that the line of sight from the viewpoint is unobstructed.


In this work, we created ground-truth annotations for both real and synthetic indoor scenes that have previously been used in a variety of applications~\cite{engel2023project,straub2024efm3dbenchmarkmeasuringprogress}.
We used egocentric datasets to ensure that this robotic task can be trained and evaluated from a first-person perspective that reflects natural human viewpoints.
In total, we annotated $250$ scenes, resulting in over $500,000$ individual annotations.
For the synthetic dataset, we selected ASE~\cite{engel2023project} due to its high-quality, visually convincing egocentric scenes, and annotated a subset of $225$ scenes.
For the real dataset, we used AEO~\cite{straub2024efm3dbenchmarkmeasuringprogress} for its diverse scenes and annotated all $25$ of its scenes.
Figure~\ref{fig:teaser} shows two examples where, given an RGB image, we annotate the VP tasks, producing a binary placement map. A person typically sits on a couch or bed (Sitting Suggestion), with the TV placed in front of it at a comfortable height (TV Placement), while the 2D Panel is best placed on a nightstand or desk (2D Panel Placement).
We also provide text descriptions of the objects within each image, along with their respective positions, to enable the use of text-based methods for VP.



Our annotations include binary masks and text descriptions in both 2D and 3D for each scene.
First, we perform 2D placement annotations separately for each VP task. 
Since the VP problem is strongly influenced by human experience, the annotations must reflect human preferences. 
For example, if a robot were to place a laptop on the floor, someone could step on it, whereas a person would naturally place it on a table.
To ensure such preferences are captured, we generated the ground truth using a combination of manual tagging and automatic tagging with a VLM (LLaVA~\cite{liu2023llava}), which was prompted to reason about human preferences.
Next, we used the available 2D bounding boxes of objects in each frame (provided in all scenes) to generate text descriptions. This was done by cropping each object and passing it individually to LLaVA~\cite{liu2023llava} to obtain its description.
Finally, to produce 3D annotations and corresponding 3D text descriptions, we reprojected the 2D points into 3D using the available pose and depth information of each frame.

We found that well-known detection and segmentation models struggle with VP tasks (e.g., IoU $\leq 0.46$ on Sitting Suggestion) when trained on our benchmark~\cite{mees20icra_placements,kirillov2023segment,keetha2023anyloc,tschannen2025siglip2multilingualvisionlanguage,ramrakhya2024seeingtheunseen,yan2018second,Lang_2019_CVPR,Liu_2024_ICRA_LPSNet,shi2019pointrcnn}.
These methods are less effective at capturing human preferences, particularly when occlusions occur in a single frame, due to their lack of global scene understanding.

Hence, our work makes the following contributions:
\begin{enumerate}
    \item We introduce Placement Aria, the first VP benchmark that addresses the placement of both small and large, natural and man-made objects. Our annotations include 2D images and 3D point clouds, with per-object text descriptions and bounding boxes for three distinct VP tasks.

    \item We provide manual annotations for all three tasks and propose an automatic tagging method, leveraging a VLM, that captures human preferences and improves the scalability of the dataset.

    \item We evaluate several models on our benchmark to establish baselines and highlight the challenges of VP.  
\end{enumerate}

\section{Related work}
\label{sec:related}


Finding a valid target placement within a scene that aligns with human preferences is non-trivial. It is required for both visual place recognition and object placement tasks.
Visual place recognition refers to the ability of systems to recognize previously visited locations based on visual input; see~\cite{7339473} for a comprehensive survey. One of its most common applications is assistive navigation for people with disabilities~\cite{Sheng_2024_NYC_Indoor_VPR}.
Our work extends egocentric data toward human-centric sitting suggestion, as well as assistive placement of common household objects.

Object placement has been extensively studied in both robotics and computer vision.
In robotics, it typically involves learning spatial relations by relying on object geometries to enable robots to perform tasks such as object placing and manipulation~\cite{jiang2012learning_place,jund2018optimization,dwibedi2017cut,jiang2012learning_arrangements, zampogiannis2015learning}. 
For example,~\cite{jiang2012learning_place} places objects into specific targets (e.g., a table), while~\cite{jund2018optimization,zampogiannis2015learning,dwibedi2017cut} arrange objects on a surface without interfering with one another.
Similarly,~\cite{jiang2012learning_arrangements} explores placements that do not obstruct people within a room.
In computer vision, object placement generally refers to inserting a real object from a source image into a target image~\cite{DBLP:journals/corr/abs-2107-01889,niu2021making,lu2023tf,zhou2022learning,lee2018contextawaresynthesis,meng2023interactiveobject,lee2018context}.
Prior works~\cite{zhou2022learning,meng2023interactiveobject} focus on determining a {\em single} appropriate location and scale for the source object, while others emphasize rendering it realistically in the target image~\cite{Zhu_2023_CVPR,lu2023tf}; see~\cite{niu2021making} for a comprehensive survey.
In contrast, our VP dataset seeks to identify all placement regions within a scene, either for placing objects or for sitting, making it a more semantic and human-centric task.

Some works~\cite{yoffe2023octopusopenvocabularycontenttracking,10.1145/3551349.3561160} proposed real-world image datasets for virtual placement. However, their sizes (100 and 308 images, respectively) make them impractical for learning at scale. In contrast, our benchmark provides over 500,000 annotations.
In~\cite{ramrakhya2024seeingtheunseen}, Ramrakhya et al. introduced a virtual placement dataset covering nine small man-made objects. Text descriptions were used to identify images containing these objects, and SAM\cite{kirillov2023segment} was applied to localize them within the images. Inpainting was then employed to remove the target objects, creating training images. Overall, their dataset consists of 10 indoor scenes and 1.3M images, which were augmented from a subset of $49,000$ images drawn from the LAION~\cite{schuhmann2022laion} and HSSD~\cite{khanna2024habitat} datasets.
However, the annotations in~\cite{ramrakhya2024seeingtheunseen} do not explicitly account for human preferences, and thus may fail to capture all plausible placement locations. For example, if a table lamp only appears on nightstands in the source images, a table would never be considered a valid placement. 
This limitation highlights the importance of our tagging strategy, which combines human expertise with VLM-guided reasoning about human preferences. 
{\bf Assistant Placement Aria} provides three semantically distinct VP tasks (2D Panel Placement, Sitting Suggestion, and TV Placement) and encodes human preferences through a combination of manual annotation and VLM-guided tagging. 
This ensures resulting plausible placements are captured not with human-centric reasoning about comfort, accessibility, and context.

In this work, we focus on egocentric indoor scene datasets, enabling a more complete understanding of human preferences. These datasets can be broadly divided into real and synthetic, and have been used to examine tasks such as segmentation, object detection, and room classification.
Real indoor datasets such as SUN-RGB-D~\cite{song2015sun}, ScanNet~\cite{dai2017scannet}, and Matterport3D~\cite{Matterport3D} provide reconstructions of large-scale indoor scenes, typically recorded using RGB-D cameras. More recent datasets, including Replica~\cite{replica19arxiv}, Aria-Twin~\cite{pan2023aria}, and AEO~\cite{straub2024efm3dbenchmarkmeasuringprogress}, offer higher reconstruction quality compared to earlier efforts. In particular, AEO provides 25 inherently different room scenes, making it both large and diverse which are the reasons that motivated our decision to annotate it.
Synthetic scene datasets, on the other hand, enable high-quality reconstructions\cite{pan2023aria}. 
Some, such as HyperSim\cite{roberts2021hypersim} and OpenRooms~\cite{li2021openrooms}, leverage online 3D models but lack real-world counterpart recordings, leaving a gap between simulated and real data. 
We chose Aria Synthetic Environments (ASE)~\cite{engel2023project} which stands out by offering both high-quality and visually convincing egocentric scenes.

\section{The Assistant Placement Aria benchmark}
\label{sec:dataset}
VP entails identifying all potential locations for placing an object within a scene. This task requires human-level semantic understanding of the environment and is particularly relevant for object placement and human assistance.
In contrast to traditional object placement and assistive navigation, which focus on a single (and usually pre-determined) destination, VP focuses on mapping out all feasible placements within the scene that align with human preferences. This enables the development of systems that can automatically select the most appropriate destination.

VP datasets are scarce, and existing works do not explicitly account for human preferences, which are essential to the task. Current datasets do not address the placement of large man-made objects or human-aiding tasks, both of which require considering global scene constraints, local physical constraints, and human preferences, e.g., placing a TV on a wall in front of a sofa for comfortable viewing.

\begin{figure*}[tb]
\centering
\begin{tabular}{cccc}
\includegraphics[width=0.2\textwidth]{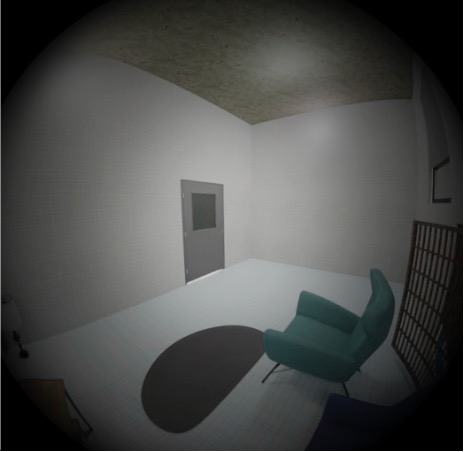} & \includegraphics[width=0.195\textwidth]{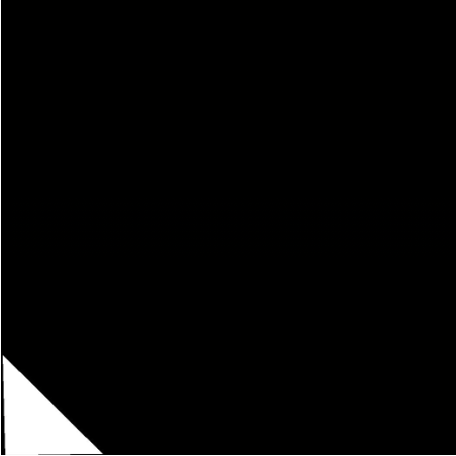} & \includegraphics[width=0.2\textwidth]{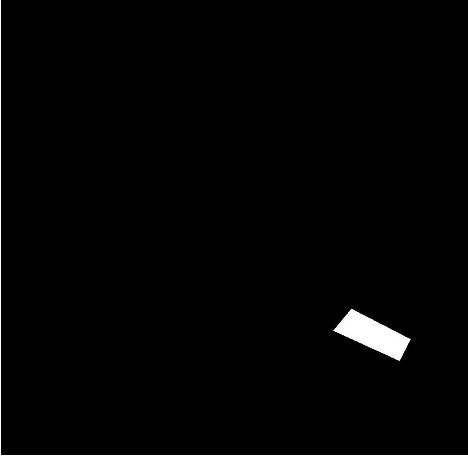} & \includegraphics[width=0.195\textwidth]{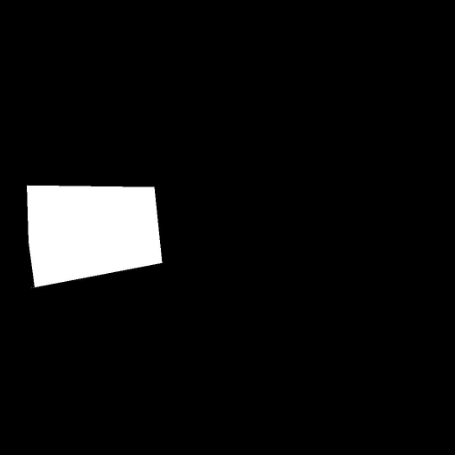} \\
(a) RGB input & (b) Panel ground truth & (c) Sitting ground truth & (d) TV ground truth \\
\end{tabular}
\caption{
{\bf Ground truth example.} 
Each frame was annotated separately for each task. (a) The original RGB image. (b) The only plausible place to place a 2D-Panel is on the desk at the bottom left corner of the image. 
(c) A person would sit on the couch. 
(d) The TV should be placed high enough over the wall and in front of the couch to enable comfortable watching.
}
\label{fig:2D placement}
\end{figure*}

To address this scarcity, we annotate a novel VP benchmark consisting of both synthetic and real data egocentric RGB-D cameras to reflect human perception:
(1) the Aria Synthetic Environments (ASE) dataset~\cite{engel2023project}, from which we selected a subset of $225$ high-quality, realistic synthetic indoor scenes; and
(2) the Aria Everyday Objects (AEO) dataset~\cite{straub2024efm3dbenchmarkmeasuringprogress}, which contains $25$ real-world scenes with a high degree of variation.
All scenes provide sequences of RGB images and 6DoF camera poses; ASE additionally includes depth images and instance segmentation maps. A 3D point cloud is available for each scene and can be further densified by reprojecting individual frames using their associated depth and pose data.
We augment both datasets in two ways: (1) by creating 2D and 3D VP annotations for each task, and (2) by generating textual descriptions of the objects in each frame, along with a global scene-level description summarizing all objects in the 3D scene, to support and enhance the annotation process.

\subsection{Tasks and Labeling}
Our annotation includes three distinct placement tasks:
{\bf (1) 2D Panel Placement}: Identifying suitable locations for placing a 2D panel representing small man-made objects, which primarily involves detecting vacant surfaces. This task is particularly relevant to object placing and manipulation, as it enables determining plausible target surfaces in a given scene. To generalize across object types, we use the abstract concept of a 2D horizontal panel (perpendicular to the $z$-axis) to represent objects ranging in size from $8$ cm in diameter (e.g., a cup) to $40$ cm (e.g., a large laptop).
{\bf (2) Sitting Suggestion}: Finding feasible sitting areas for a person, which is important for aiding people with disabilities. 
This task particularly involves various human-centric constraints such as adequate space for different body parts, along with considerations of comfort and ergonomics.
{\bf (3) TV Placement}: Determining appropriate positions for a TV screen, which is relevant to human-assistance tasks. This task requires accounting for both global and local factors, including optimal viewing height, unobstructed wall space, and ergonomic viewing angles. 
For example, the TV should not be mounted in the top or bottom $20\%$ of the wall height to enable comfortable viewing, and should support sizes ranging from $43$ to $65$ inches (typical TV sizes).
These tasks span a wide range of object sizes and spatial reasoning challenges, each reflecting different complexities to VP.

As the object placement task is strongly influenced by human experience, we manually annotated $50$ scenes. However, manual labeling limits scalability to new tasks. To address this, we also introduce an automatic labeling approach, which we applied to an additional $200$ scenes.
All annotations are conducted separately for each task in 2D using a binary classification scheme: placeable regions are labeled as {\it 1}, and non-placeable regions as {\it 0}. Fig.~\ref{fig:2D placement} shows a typical example of our annotations across the three placement tasks.
Both manual and automatic 2D annotations are then reprojected into 3D to generate a placement point cloud for each scene using the available camera poses. 
The 3D annotation follows the same binary scheme based on their suitability for the given task, i.e., point in placeable regions are labeled as {\it 1}, and non-placeable regions as {\it 0}.

\begin{figure*}[tb]
\centering
\begin{tabular}{c@{\hspace{2pt}}c@{\hspace{2pt}}c@{\hspace{2pt}}c@{\hspace{2pt}}c@{\hspace{8pt}}c@{\hspace{8pt}}c}
   
 \rotatebox{90}{ 2D-Panel} &
   \includegraphics[width=0.15\textwidth]{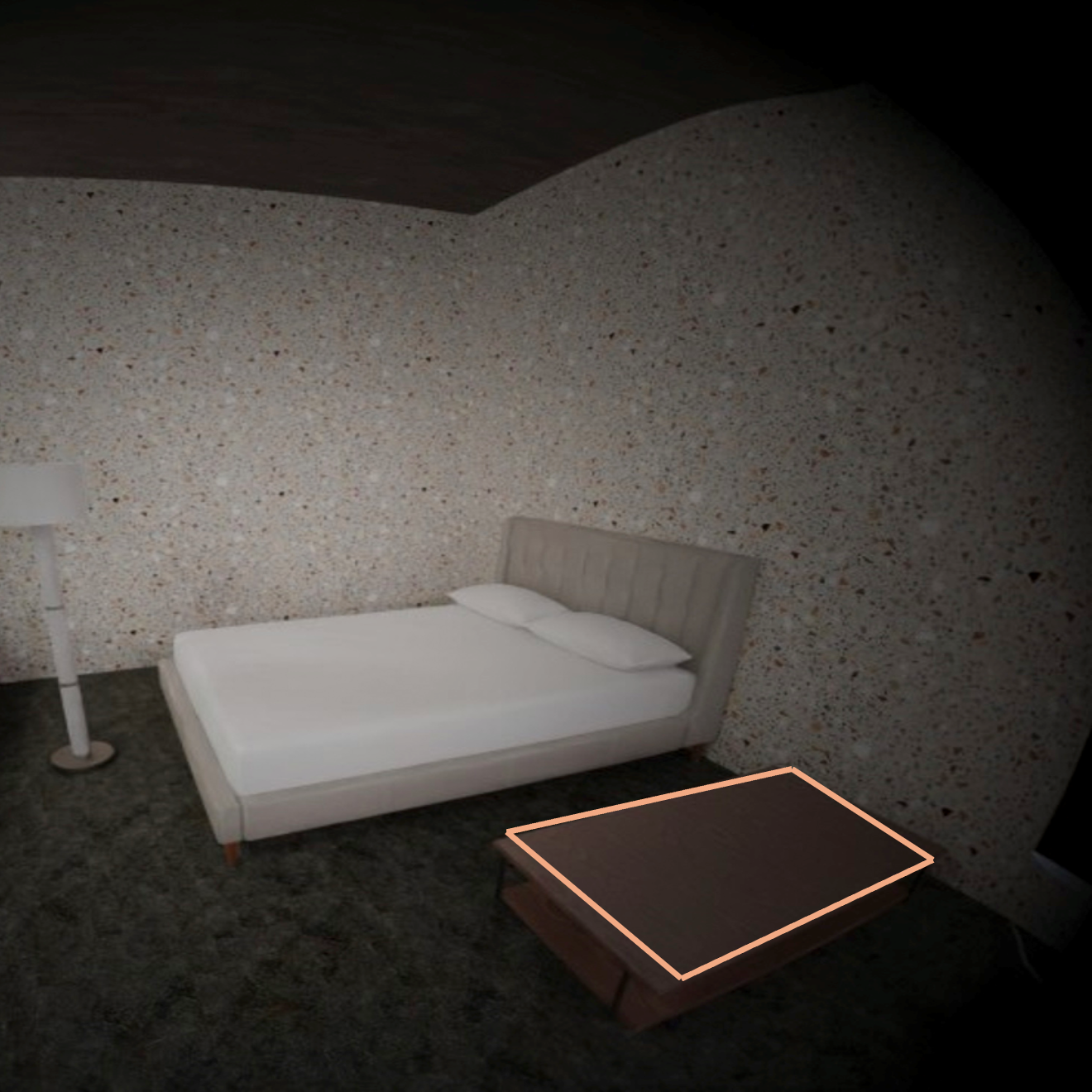} & 
   \includegraphics[width=0.15\textwidth]{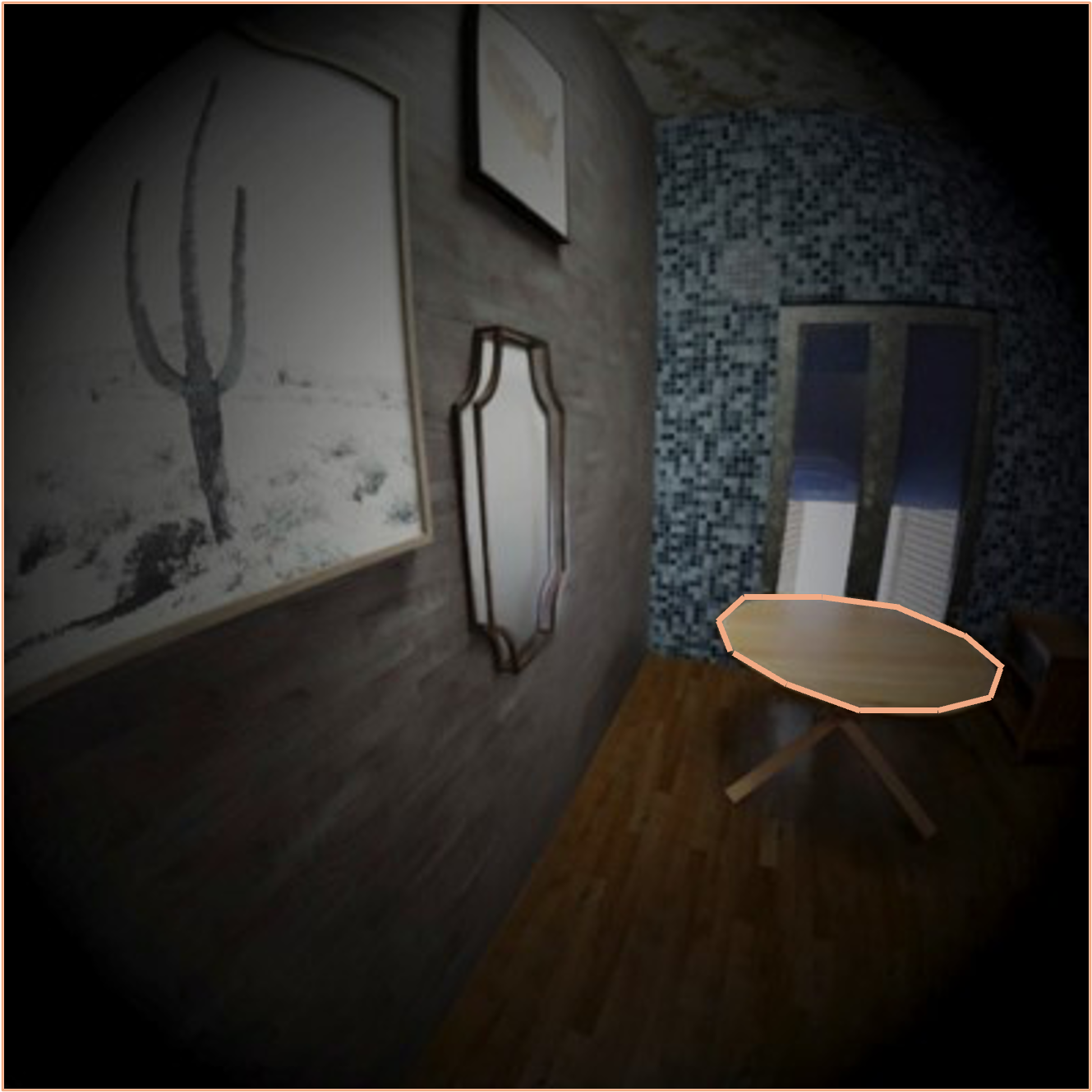} &
   \includegraphics[width=0.15\textwidth]{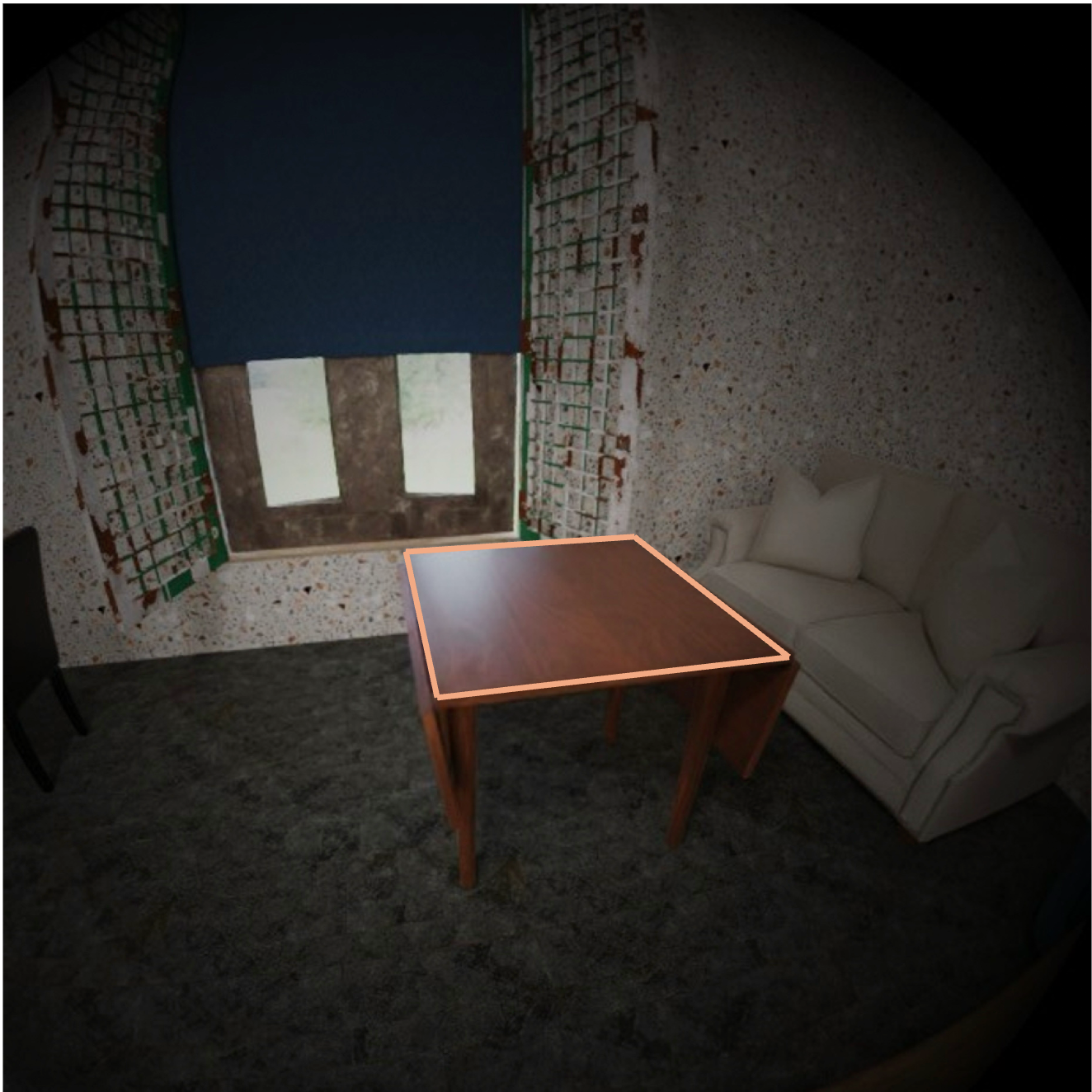} &
   \includegraphics[width=0.15\textwidth]{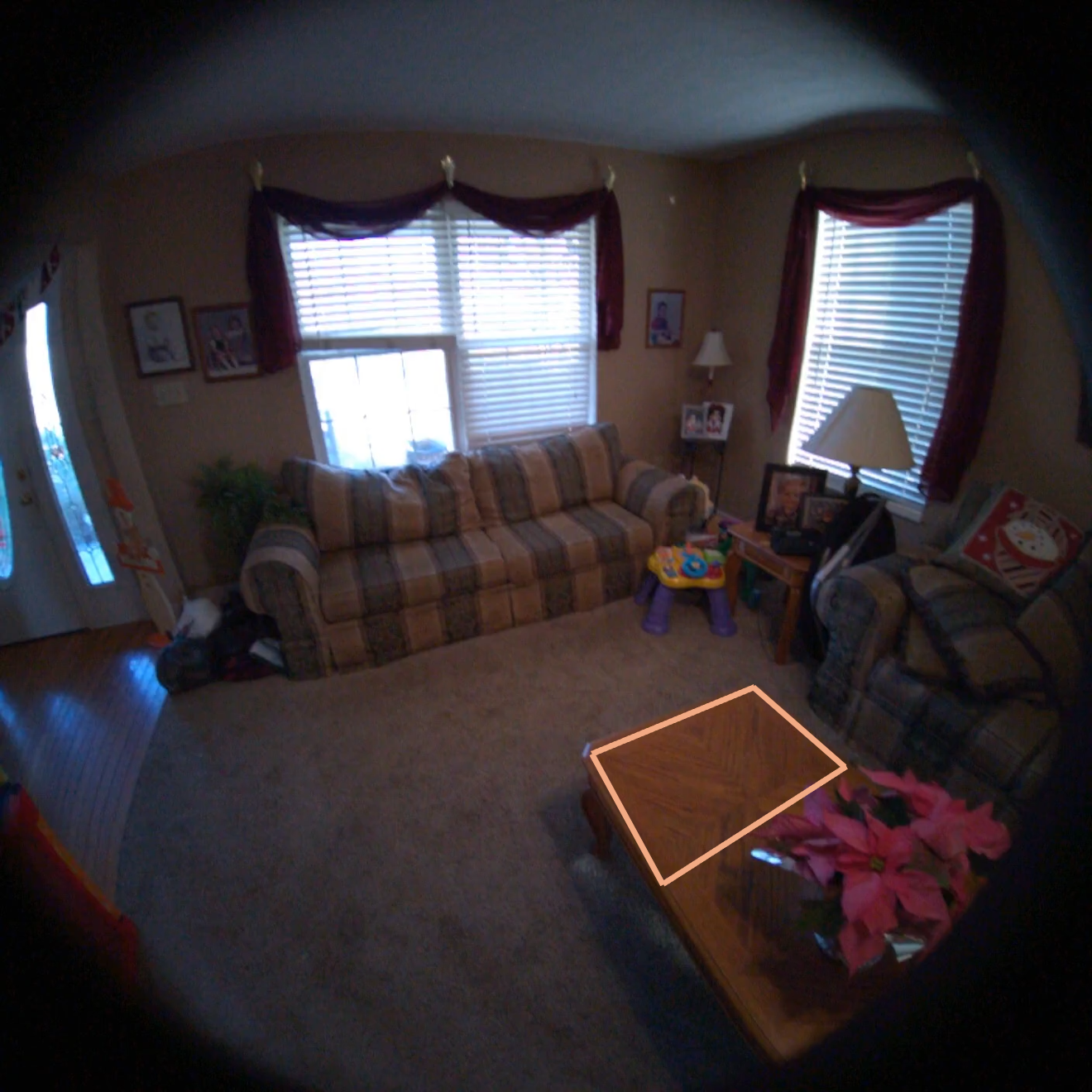} &
   \includegraphics[width=0.15\textwidth]{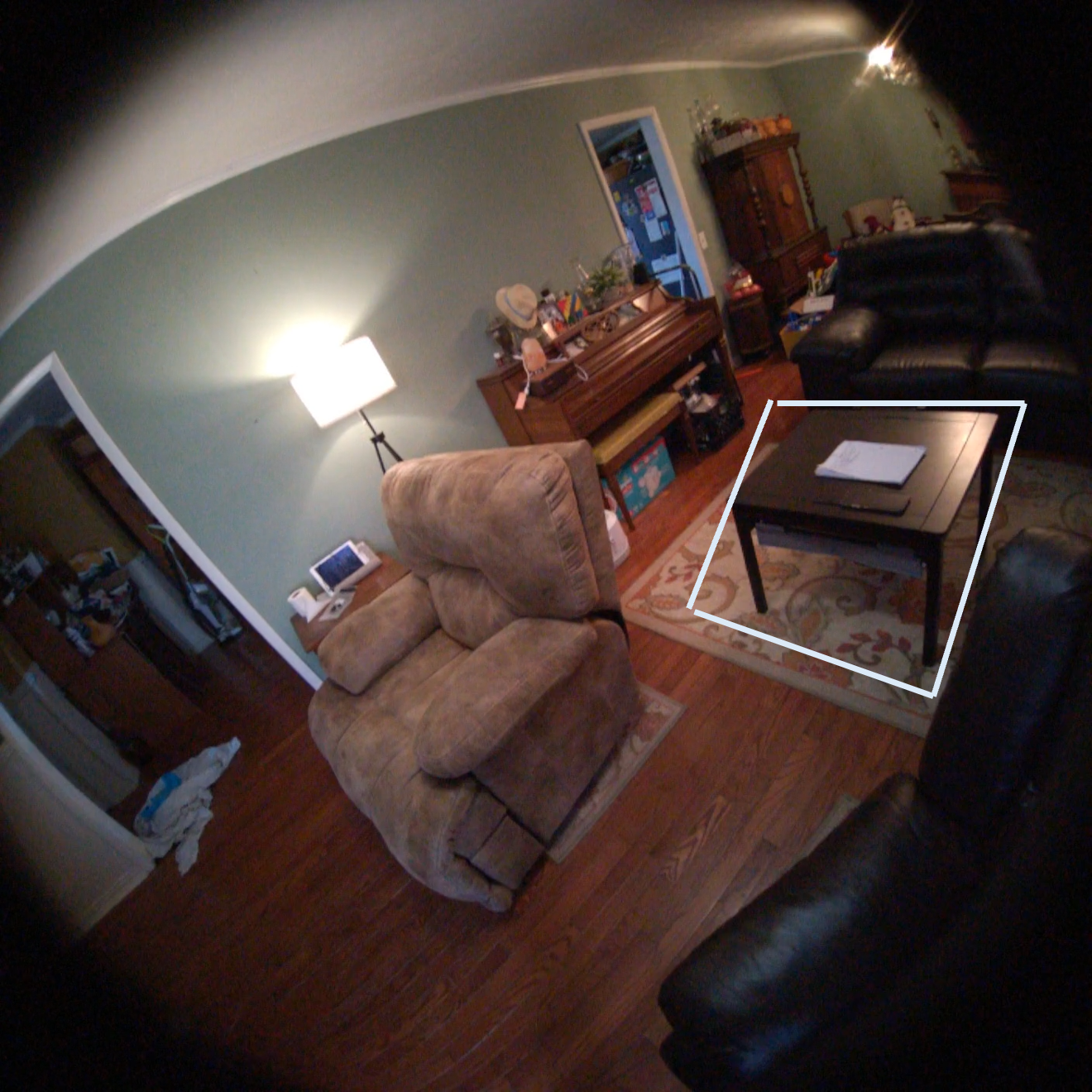} & 
   \includegraphics[width=0.15\textwidth]{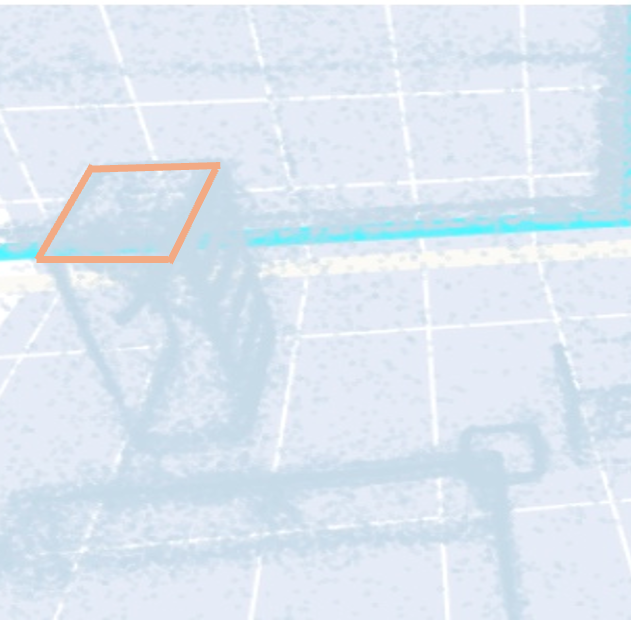} \\ 
 
 \rotatebox{90}{Sitting} &
  \includegraphics[width=0.15\textwidth]{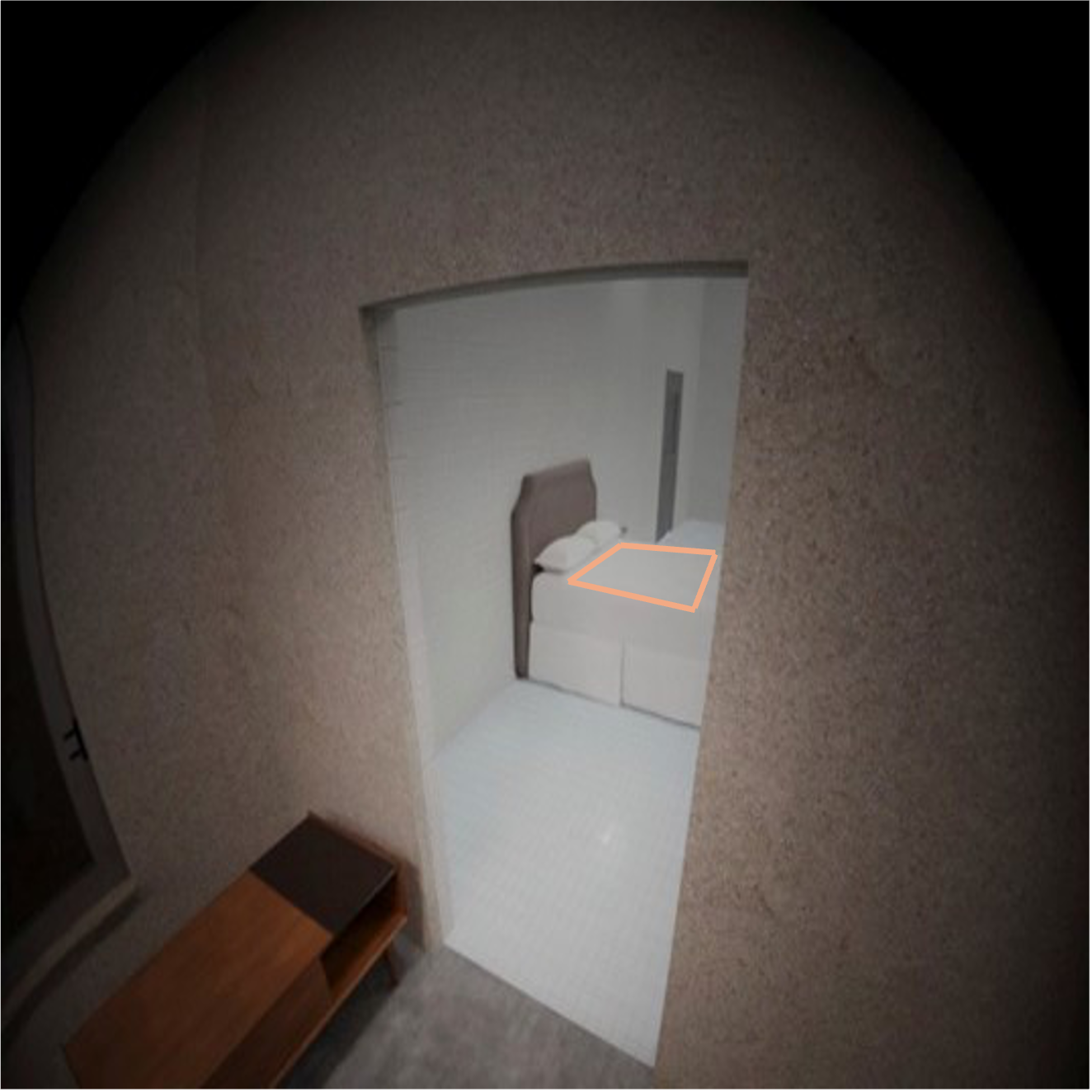} &
  \includegraphics[width=0.15\textwidth]{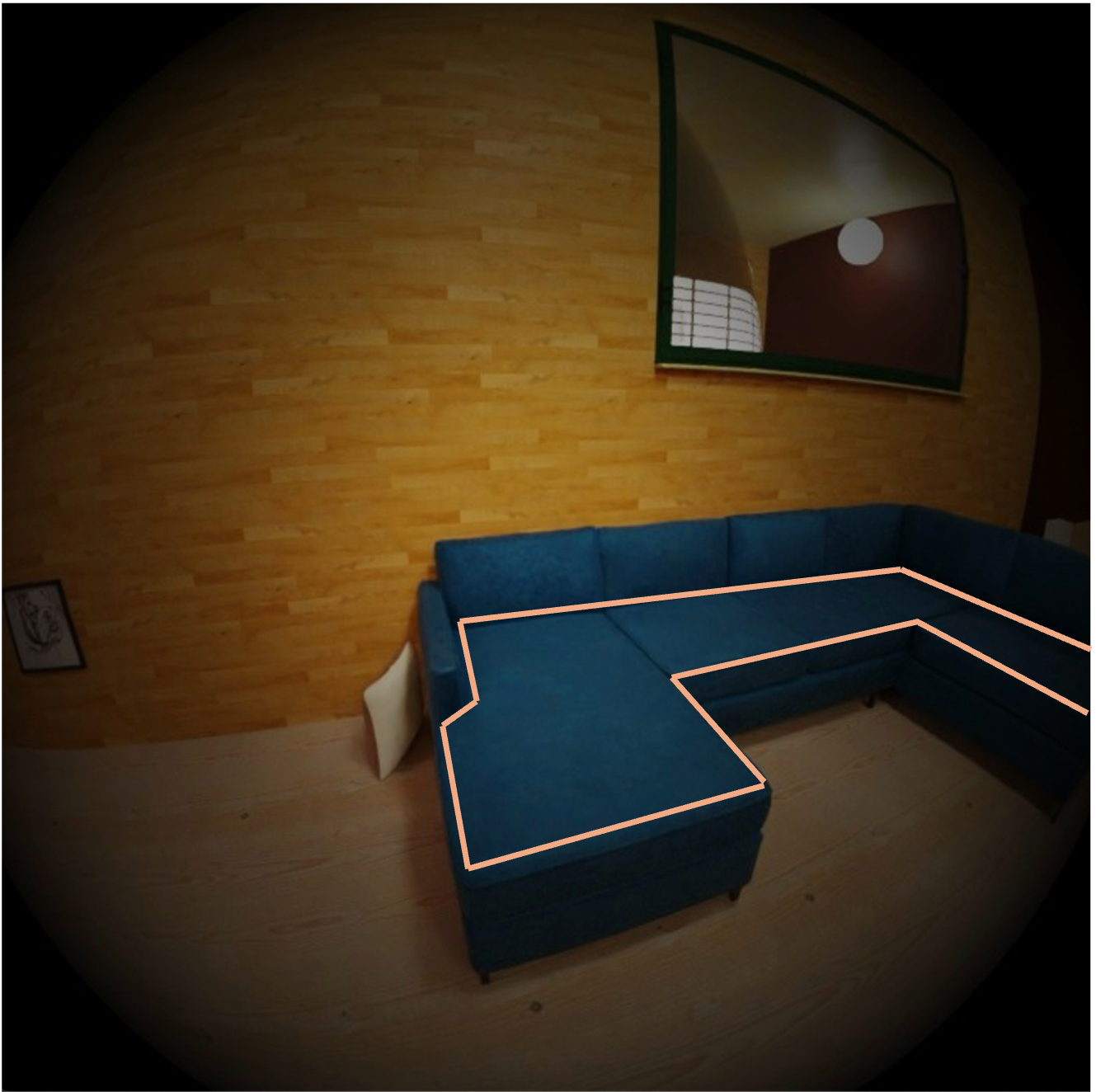} &
  \includegraphics[width=0.15\textwidth]{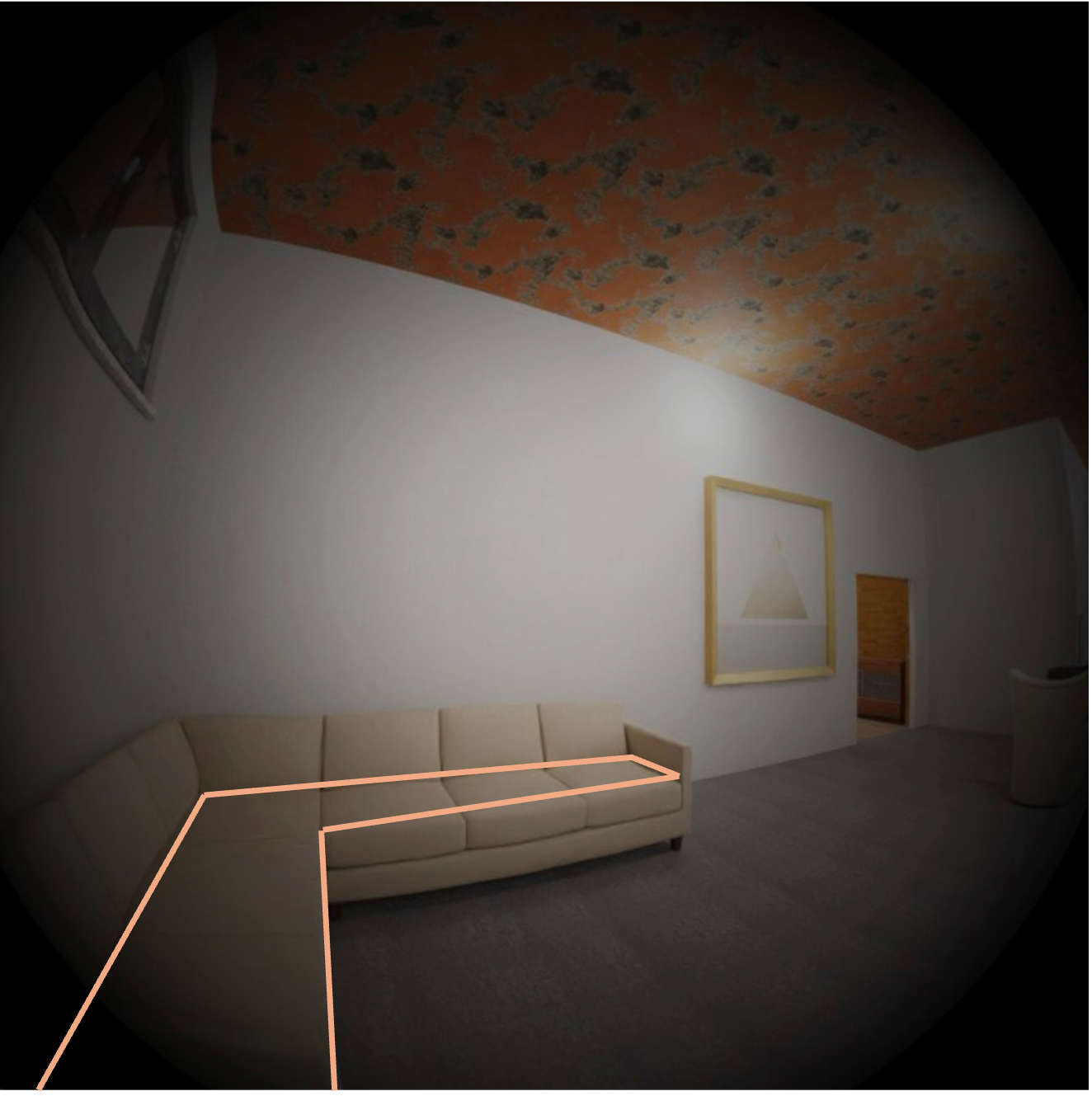} &
  \includegraphics[width=0.15\textwidth]{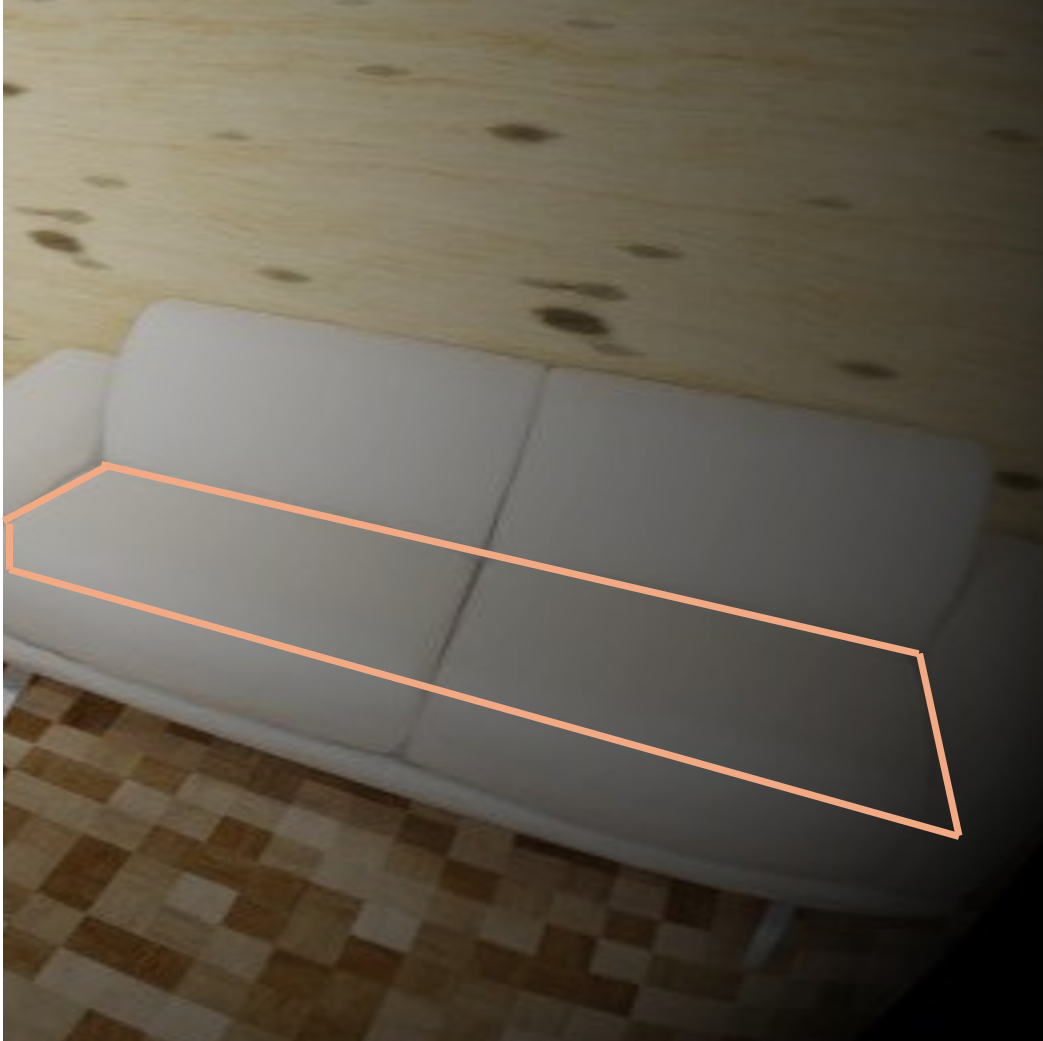} &
  \includegraphics[width=0.15\textwidth]{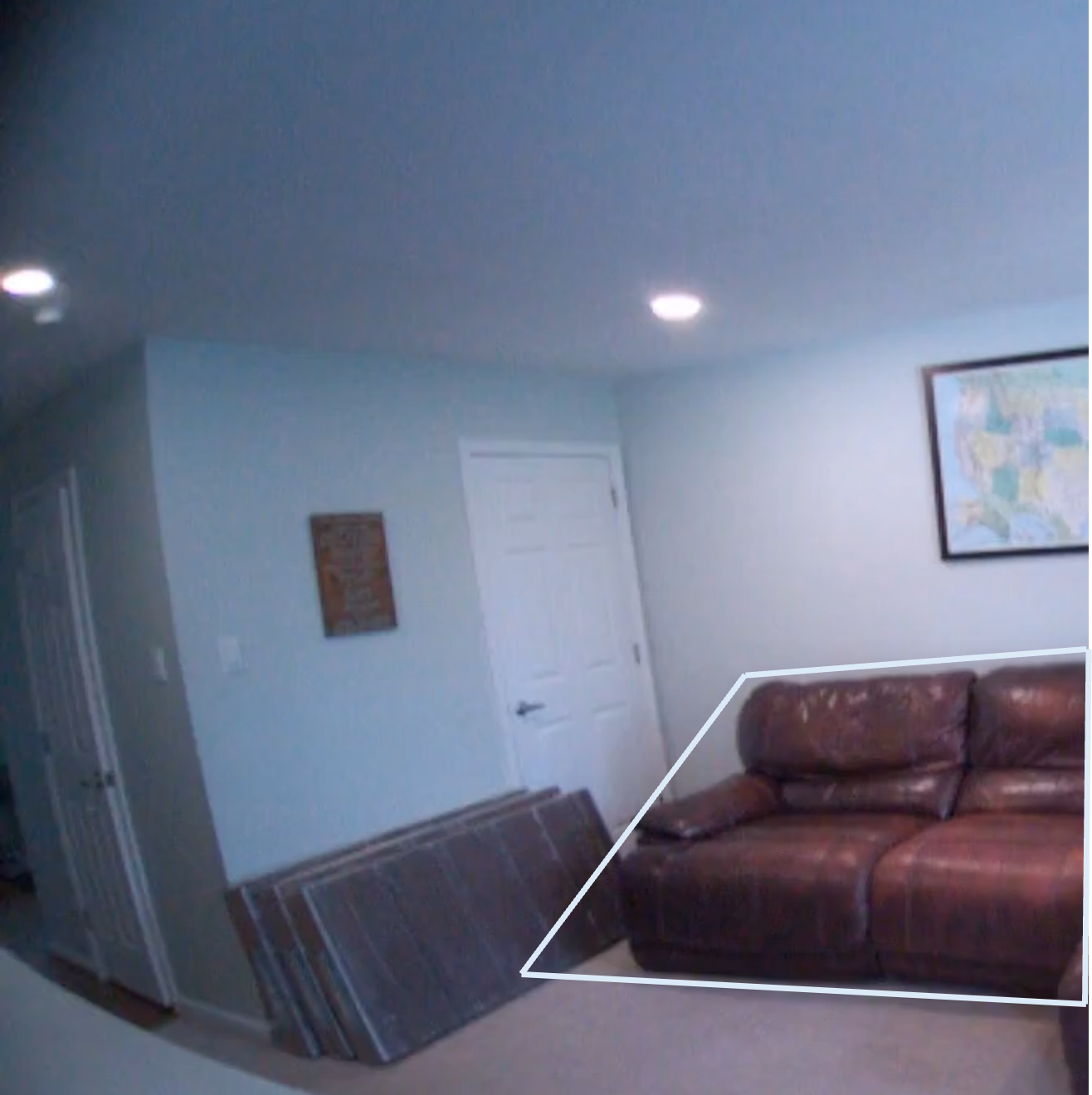} &
  \includegraphics[width=0.15\textwidth]{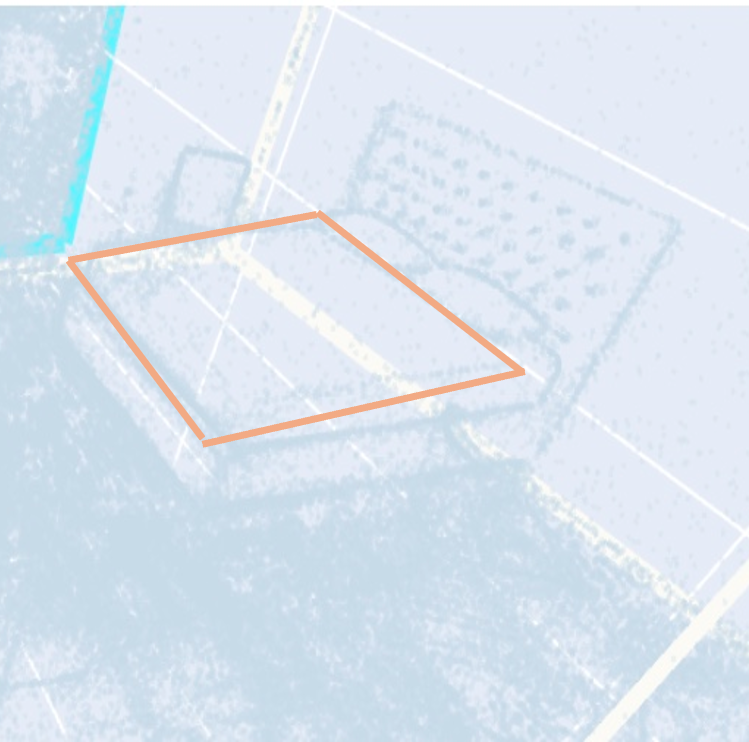} \\

 \rotatebox{90}{TV} &
   \includegraphics[width=0.15\textwidth]{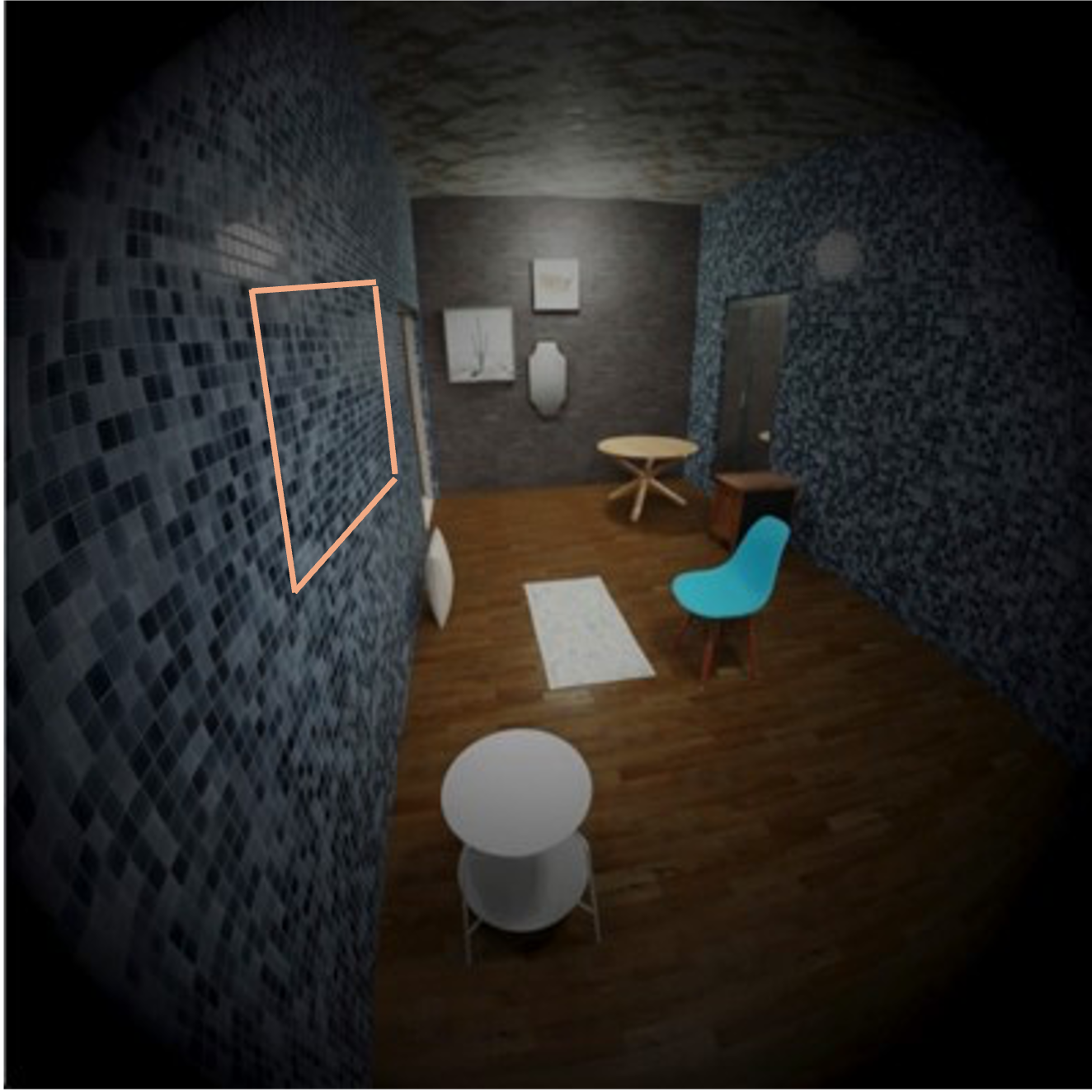}&
   \includegraphics[width=0.15\textwidth]{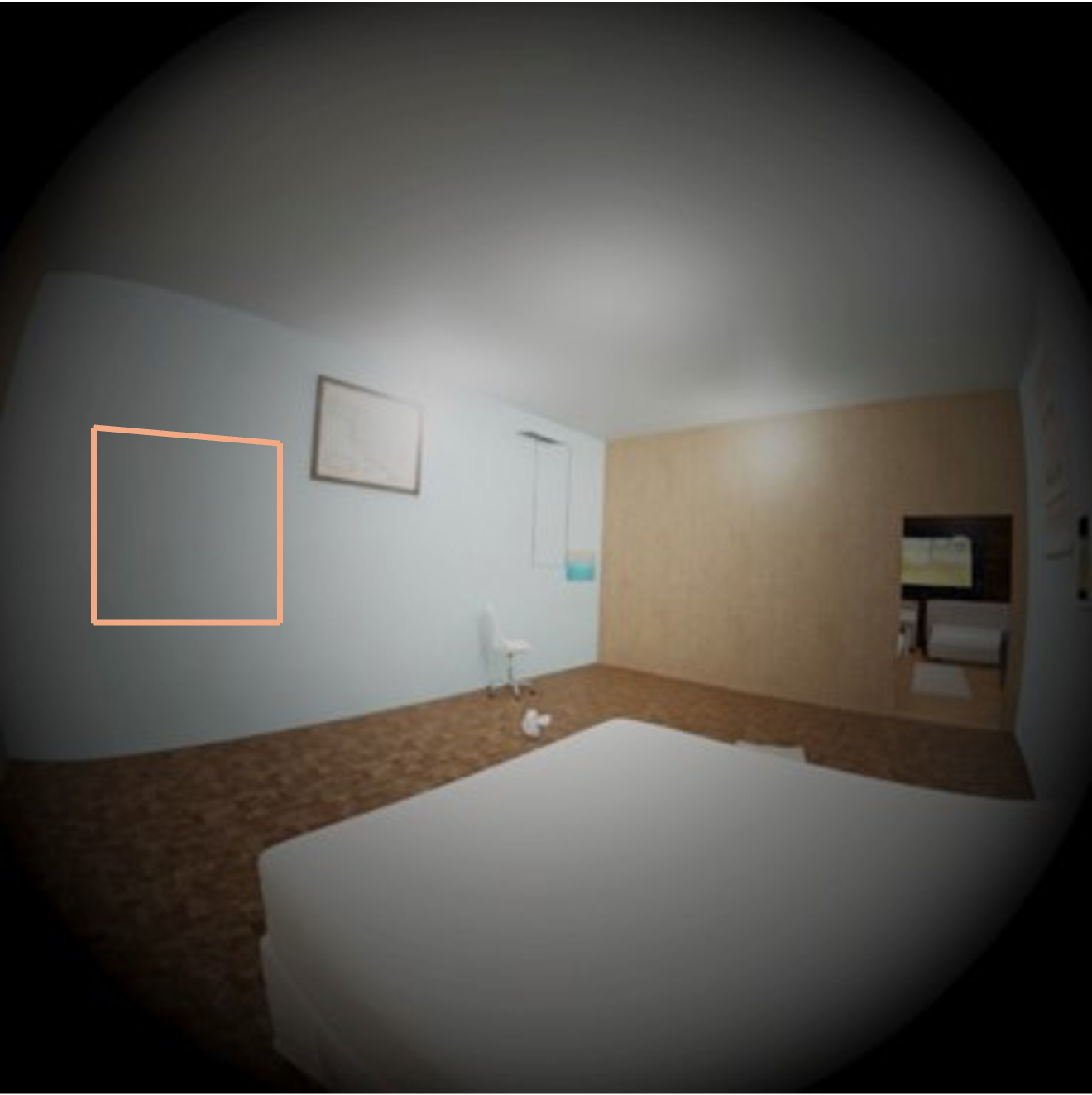}&
   \includegraphics[width=0.15\textwidth]{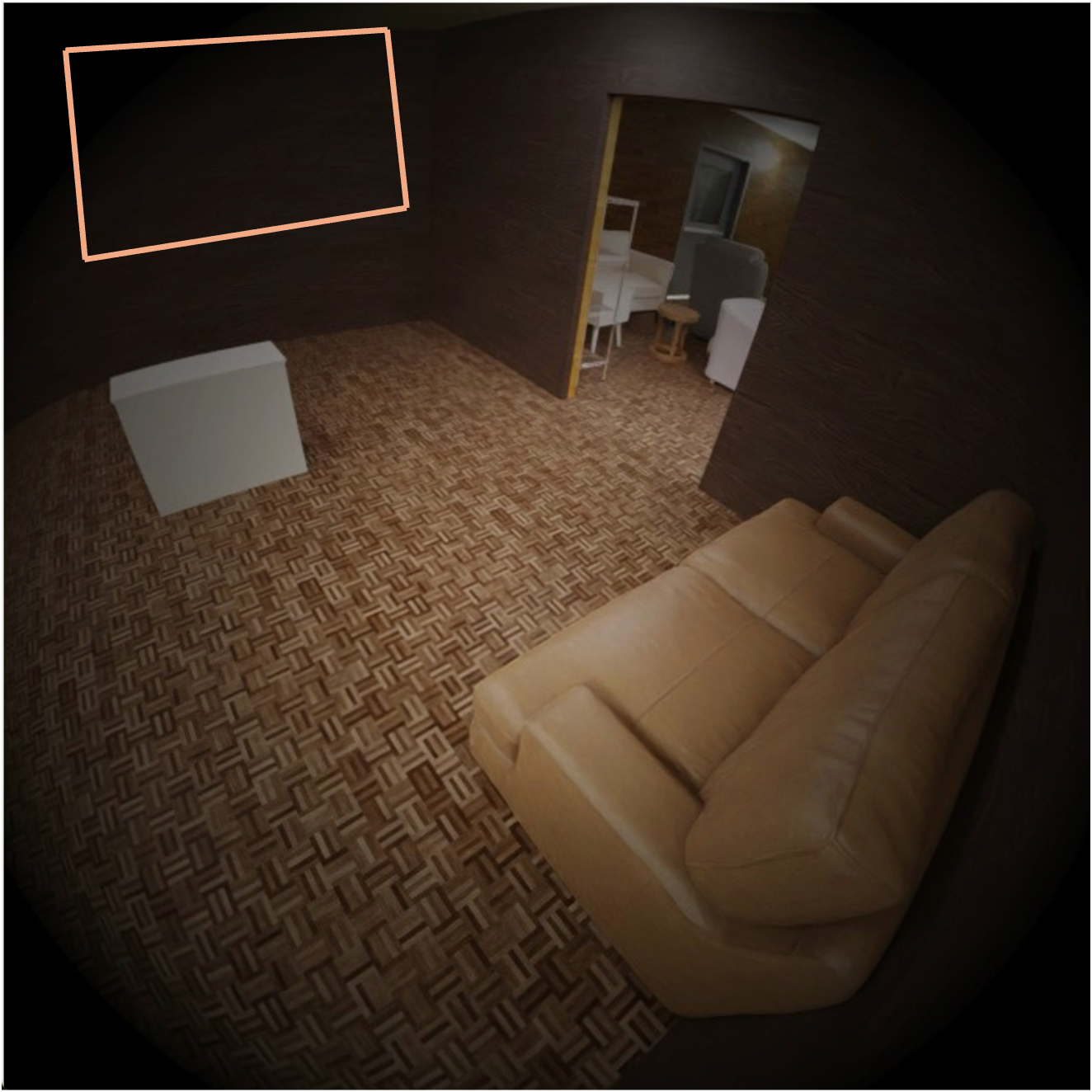}&
   \includegraphics[width=0.15\textwidth]{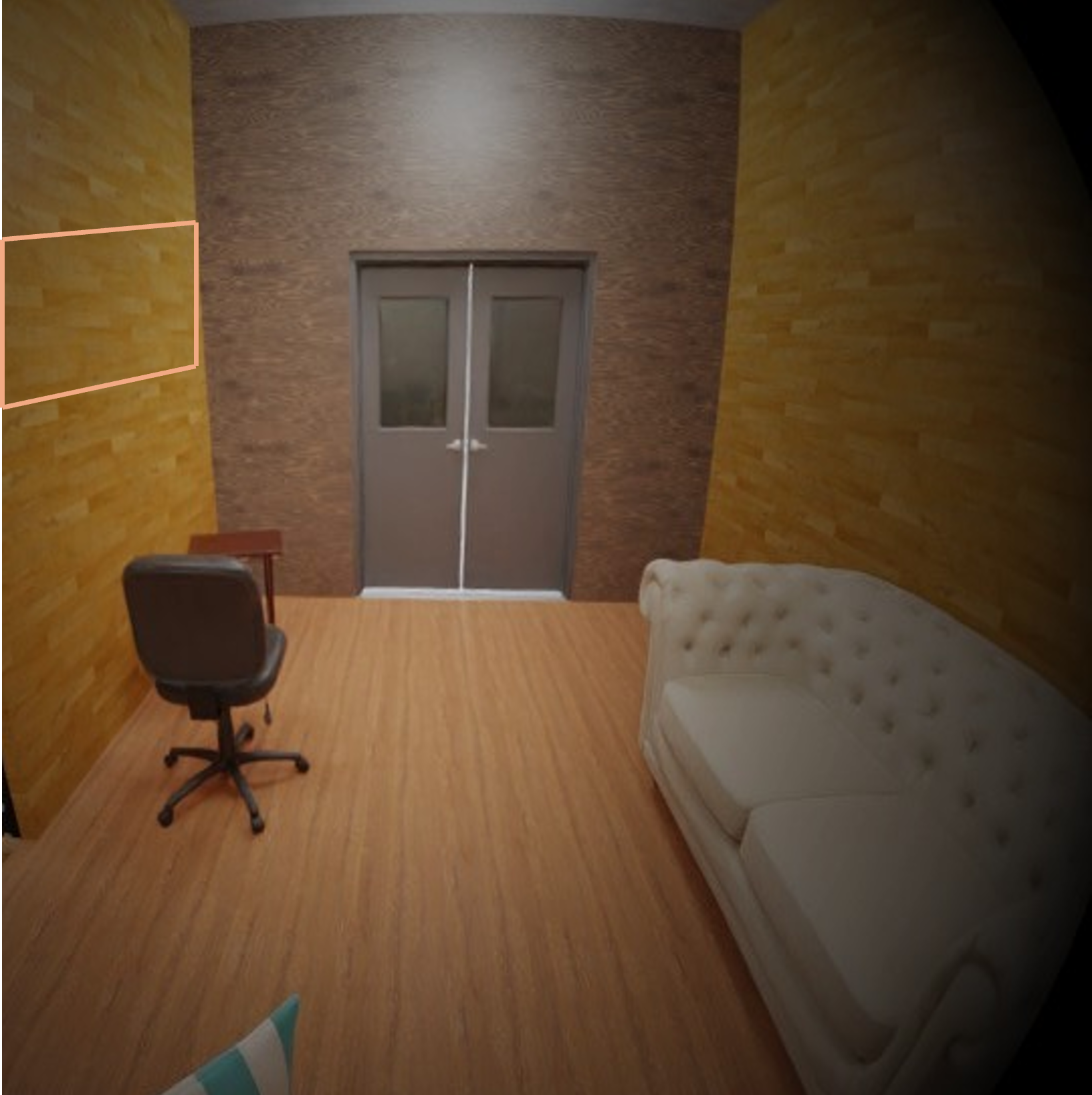}&
   \includegraphics[width=0.15\textwidth]{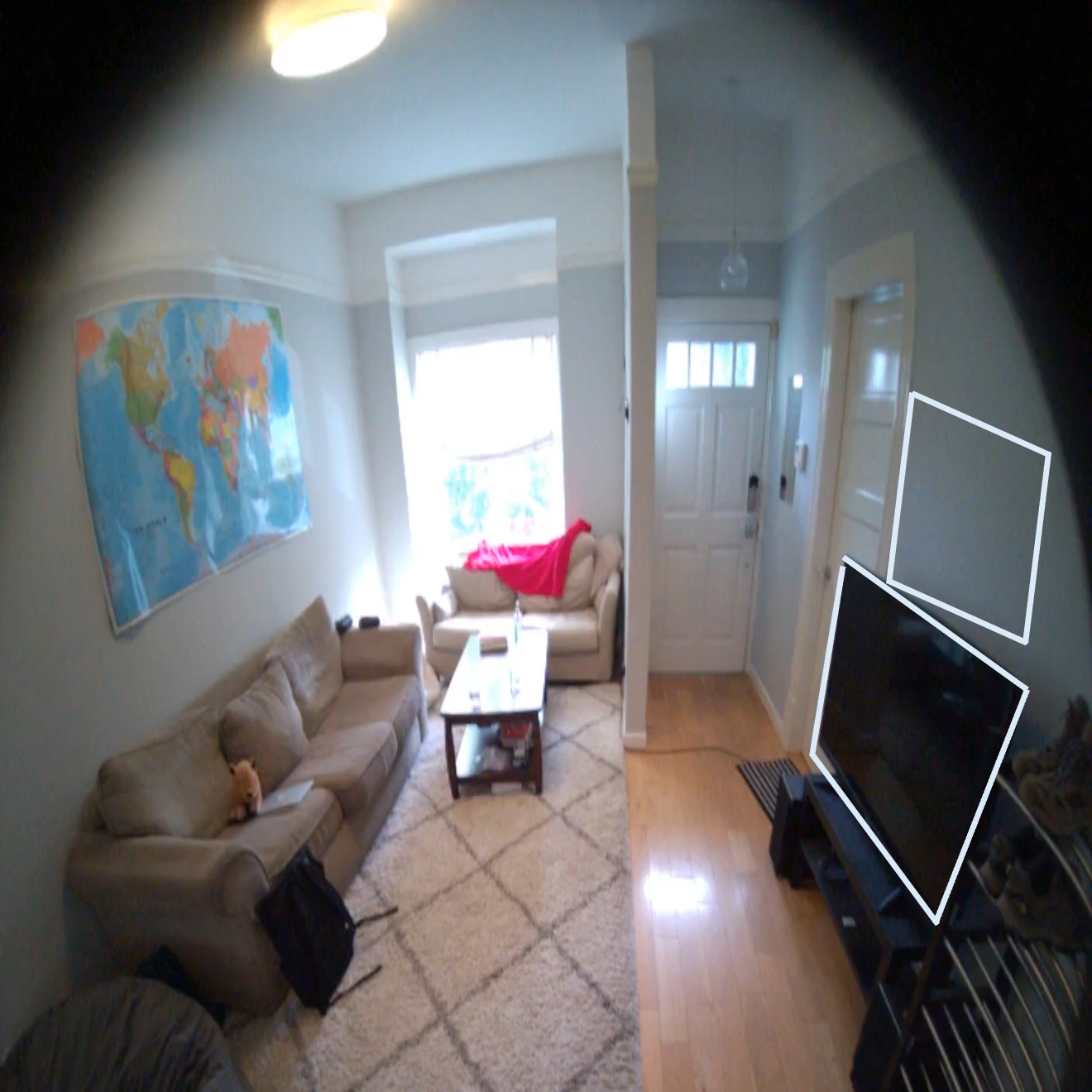}&
   \includegraphics[width=0.15\textwidth]{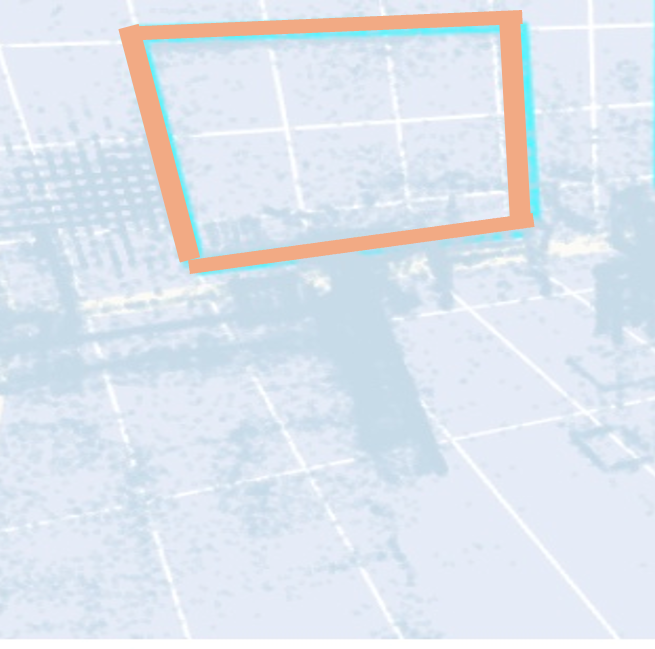}\\
  
& \multicolumn{4}{c}{(a) Manual} & (b) Auto. & (c) PC \\
\end{tabular}
\caption{
{\bf Annotation examples.} 
Each task is shown with $4$ manually annotated frames (a), followed by one automatically generated annotation (b), and one 3D point cloud annotation (c) reprojected from the annotated frames.
The manual annotations highlight the high quality and precision of the human tagging. 
The automatic tagging is a bit less precise, as it covers entire objects.
}
\label{fig:2D placement examples}
\end{figure*}

{\bf Manual labeling.}
The annotations were performed by a team of three expert human taggers. Each task was defined as clearly and simply as possible to minimize bias during tagging and was subject to the constraints outlined above. 
For example: {\em "Where would one place a TV screen whose size is between $43$ and $65$ inches?"}.
Fig.~\ref{fig:2D placement examples}(a) shows several 2D examples from each task, while Fig.~\ref{fig:2D placement examples}(c) illustrates several 3D reprojections. In cases of disagreement over a specific area or pixel, the majority vote was taken. For instance, one such disagreement was whether a single chair should be considered a comfortable place to view a TV from (the majority voted {\em yes}).
It is worth noting that disagreements were rare, occurring in fewer than $2\%$ of the annotations.

\begin{figure*}[tb]
\centering
\begin{tabular}{cc}
\includegraphics[height=0.17\textwidth]{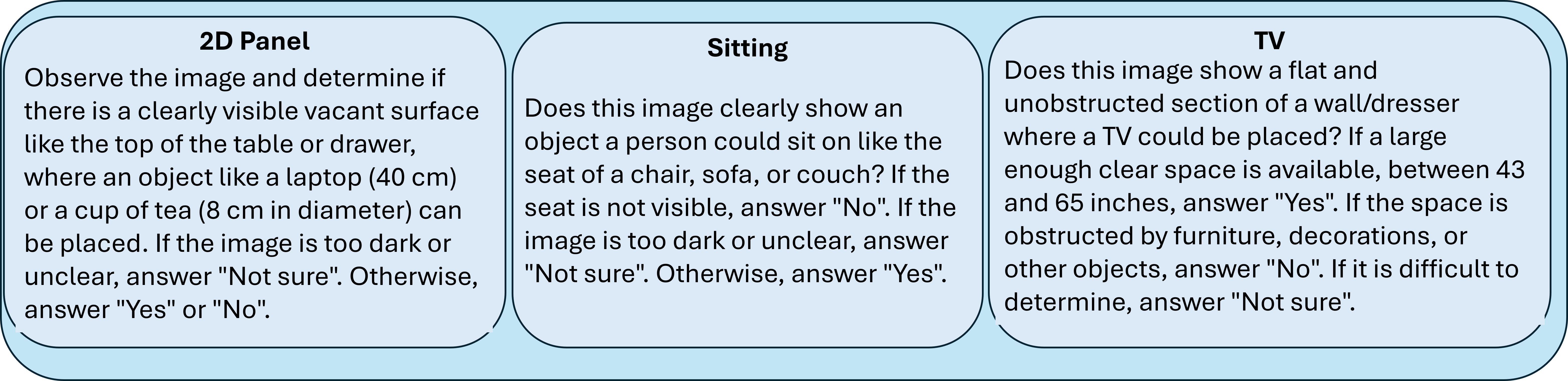} &
\includegraphics[height=0.17\textwidth]{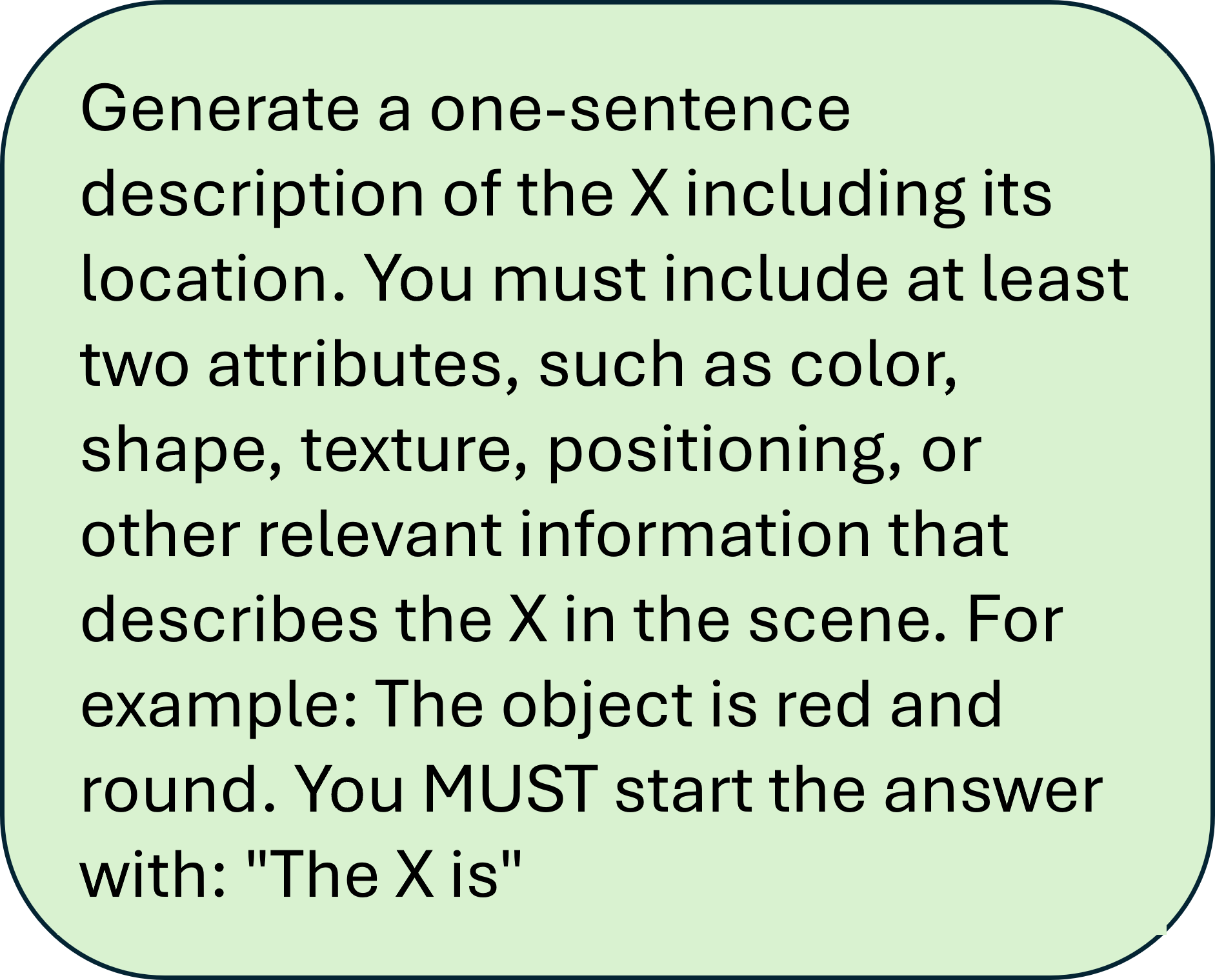} \\
(a) Assistance placement questions & (b) Text description
 \\
\end{tabular}
\caption{
{\bf Example VP task prompts and scene description.} 
The questions posed to LLaVA for the VP annotations and the text description of each frame. 
}
\label{fig:prompts}
\end{figure*}

{\bf Automatic labeling.}
As explained above, VP requires reasoning about human preferences. 
To support this, we use LLaVA~\cite{liu2023llava} for automatic annotation, leveraging its reasoning capabilities, which are known to align well with human judgment.
To guide LLaVA, it was prompted with $10$ detailed examples for each VP task, helping it to better capture human constraints. 
We annotated $200$ scenes from ASE~\cite{engel2023project}, where each scene consists of multiple frames, and each frame was annotated independently. 
ASE provides a segmentation map for every frame, with each object represented by a unique mask. 
For each object, we crop the corresponding region from the frame and input it into LLaVA. 
Then, LLaVA's labeling process (of all three tasks) begins with the following instruction: \textit{"Please answer the following questions while considering what a person would answer"}, to account for human preferences. 
This instruction is followed by a task-specific question about the object (e.g., \textit{"Could a TV screen between $43$ and $65$ inches be placed here?"}). 
If LLaVA answers \textit{"yes"}, we label the corresponding object mask as placeable; otherwise, it is labeled as non-placeable.
The full formulations of all three VP tasks are shown in Fig.~\ref{fig:prompts}(a). 
Furthermore, we utilized the 6DoF information of each scene to verify that the TV screens were not placed in the top or bottom $20\%$ of the each room.

Although useful for scaling the labeling process, the automatic procedure is less accurate than manual annotation. Automatic labeling tends to be more general in its selection of objects, rather than focusing solely on valid placement regions as a human annotator would. For example, when placing a 2D panel (e.g., a laptop), a human annotator would typically label only the tabletop, whereas the automatic annotation might also include the table’s legs, since they fall within the object mask (see Fig.~\ref{fig:2D placement examples}(b)).
To assess the accuracy of the automatic annotations, we compared them with the manual annotations on ASE’s $25$ manually tagged scenes using the Intersection-over-Union (IoU) between corresponding masks. TV Placement showed the highest agreement, with a mean IoU of $0.70$. The 2D Panel task followed with a mean IoU of $0.64$, while Sitting Suggestion was the least accurate, with a mean IoU of $0.60$, as entire furniture items were often labeled instead of just their seats.
Moreover, our human taggers manually verified that LLaVA tagged only reasonable objects for each task (e.g., chairs and couches for Sitting Suggestion), which were removed only in very rare cases (fewer than $3\%$) across all scenes. 
These failure cases typically involved partially visible objects that were not clearly captured within the frame, such as the desk in Fig.~\ref{fig:2D placement}(a).
While the automatic tagging is less accurate than manual annotation, it is worth noting that many object-placement systems only need to be directed to the target destination (e.g., Sitting Suggestion, which showed lower IoU) and can refine the available spots on a surface as part of their task definition. Thus, coarse annotations may still provide sufficient guidance for practical human-assistance tasks.

{\bf Text descriptions.}  
Since textual descriptions are valuable for a range of scene understanding tasks, we also employ LLaVA to generate descriptions for each object in every frame. 
Leveraging these descriptions may improve VP performance in text-aware systems. 
For each object in a frame, we crop the image around the object as described above. 
The cropped region is then passed to LLaVA, which is prompted to generate a description containing at least two attributes, such as color, shape, texture, or other relevant characteristics (see Fig.~\ref{fig:teaser}). 
The full prompt used to create these descriptions is shown in Fig.~\ref{fig:prompts}(b), where \texttt{X} denotes the object in question. 
This process produces a caption for each frame, consisting of the 2D bounding boxes (bboxes) and the descriptions of all objects within it. 
In addition, we generate a global caption for the entire scene, which includes descriptions of all objects together with their corresponding 3D bboxes. 
Further details can be found in the Appendix~\ref{sec:appendix}.

 {}

\begin{figure*}[tb]
  \centering
  \begin{subfigure}[t]{0.33\textheight}
    \centering
    \includegraphics[width=\linewidth]{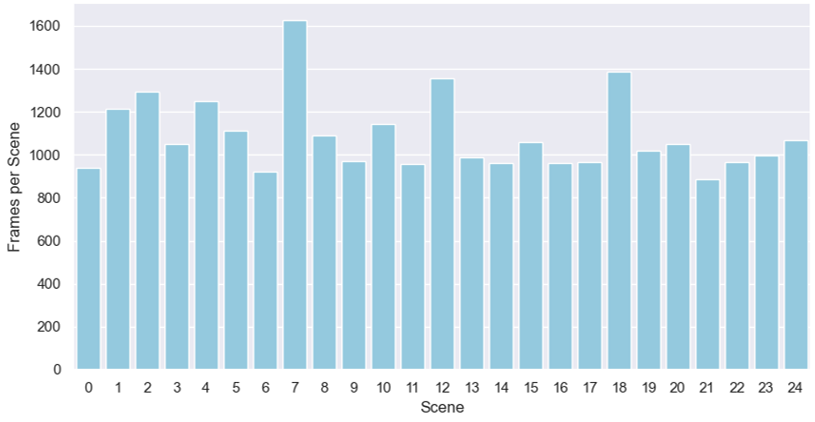}
    \caption{Real}
    \label{fig5:real}
  \end{subfigure}
  \qquad
  \begin{subfigure}[t]{0.33\textheight}
    \centering
    \includegraphics[width=\linewidth]{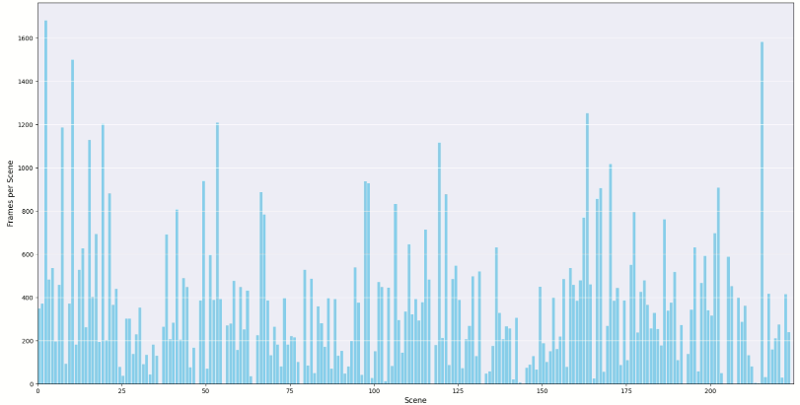}
    \caption{Synthetic}
    \label{fig5:synthetic}
  \end{subfigure}
\caption{
{\bf Number of frames per scene.} 
The number of frames within each scene.
}
\label{fig:statistics}
\end{figure*}

\subsection{Dataset Statistics and Benchmark}

As each of our scenes is unique, they vary in terms of eligible placement area, number of frames, and number of points. Figure~\ref{fig:statistics} shows the distribution of frames per scene, divided into real (a) and synthetic (b).
For the synthetic data~\cite{engel2023project}, the number of frames ranges $93$ and $1766$ with $670$ frames per scene on average. The number of points ranges from $85$ and $2637$ (thousand) with an average of $677$ thousand points per scene.
For the real data~\cite{straub2024efm3dbenchmarkmeasuringprogress}, the number of frames ranges from $920$ and $1625$ with $1100$ frames per scene on average. The number of points ranges from $289$ and $1657$ thousand with an average of $647$ thousand points per scene.

The percentage of eligible placement areas varies across tasks and data types. For 2D Panel Placement, the average percentage of pixels eligible for placement was $1.8\%$ in the synthetic data and $2\%$ in the real data. For Sitting Suggestion, $2.7\%$ of the synthetic data and $3\%$ of the real data were eligible. For TV Placement, $4.1\%$ of the synthetic data and only $1\%$ of the real data were eligible.
This difference is not surprising, as the real data contains more images of walls, leaving less vacant space for televisions. Overall, the average number of annotations per scene is $1905$, with each frame containing an average of $2.8$ annotations.

Our benchmark is divided into training, validation and test sets:
{\bf Training set}: $200$ scenes, consisting of $15$ real (scenes $10$-$24$), $15$ manually labeled synthetic (scenes $5$-$9$), and $170$ automatically labeled synthetic (scenes $40$-$209$);
{\bf Validation set}: $25$ scenes, consisting of $5$ real (scenes $5$-$9$), $5$ manually labeled synthetic (scenes $10$-$24$), and $15$ automatically labeled synthetic (scenes $25$-$39$); and 
{\bf Test set}: $25$ scenes, consisting of $5$ real (scenes $0$-$4$), $5$ manually labeled synthetic (scenes $0$-$4$), and $15$ automatically labeled synthetic (scenes $210$-$224$).

\begin{table*}[tb]
\begin{center}
\caption{
{\bf Virtual Placement results.} VP results evaluated using IoU on the Assistant Placement Aria benchmark.}
\begin{tabular}{|l|l|c|c|c|c|c|c|}
\hline
\multirow{2}{*}{Method} & \multirow{2}{*}{Input} & \multicolumn{3}{c|}{Manually tagged scenes} & \multicolumn{3}{c|}{All scenes}  \\ 
\cline{3-8}
 && \scriptsize  2D-Panel &\scriptsize Sitting &\scriptsize TV & \scriptsize  2D-Panel &\scriptsize Sitting &\scriptsize TV  \\ 
\hline
AIS~\cite{mees20icra_placements} & \multirow{5}{*}{Images} & $25.7\%$  & $29.5\%$ &  $24.1\%$  & $65.7\%$  & $59.5\%$ &  $54.1\%$    \\
 SAM~\cite{kirillov2023segment} && $35.1\%$  & $40.2\%$ & $20.7\%$ &  $70.7\%$  & $59.5\%$ &  $50.1\%$  \\
 AnyLoc-VLAD DINOv2~\cite{keetha2023anyloc} && $22.4\%$  & $24.9\%$  & $21.4\%$ &  $59.4\%$  & $57.1\%$ &  $54.0\%$  \\
 SigLIP-$2$~\cite{tschannen2025siglip2multilingualvisionlanguage} && $38.2\%$  & $39.1\%$ & $25.6\%$ &  $68.8\%$  & $60.1\%$ &  $55.9\%$  \\
CLIP-UNet~\cite{ramrakhya2024seeingtheunseen} && $\mathbf{42.7\%}$  & $\mathbf{45.7}\%$ & $\mathbf{30.7}\%$ &  $\mathbf{75.7\%}$  & $\mathbf{62.5}\%$ &  $\mathbf{60.3}\%$   \\
 \hline
 \hline
SECOND~\cite{yan2018second} & \multirow{4}{*}{Point clouds} & $15.7\%$  & $23.4\%$ & $25.7\%$ & $35.7\%$  & $30.4\%$ &  $30.2\%$   \\
PointPillars~\cite{Lang_2019_CVPR} & & $14.8\%$  & $23.1\%$ & $27.2\%$ & $24.3\%$  & $29.2\%$ &  $32.1\%$   \\
LPS-Net~\cite{Liu_2024_ICRA_LPSNet} & & $\mathbf{17.9\%}$ & $\mathbf{27.7}\%$  & $\mathbf{29.8}\%$ &  $\mathbf{39.7\%}$  & $\mathbf{34.6\%}$ &  $\mathbf{39.2\%}$     \\
PointRCNN~\cite{shi2019pointrcnn} && $17.2\%$ & $25.7\%$  & $28.7\%$ &  $38.6\%$  &  $32.4\%$ &  $32.8\%$\\
 \hline
\end{tabular}
\label{tbl: full res}
\end{center}
\end{table*}

\section{Experiments}
\label{sec:experiments}

We evaluate our dataset using IoU to compare the predicted placements with the ground-truth annotations. As baselines, we assess a range of 2D and 3D methods, which are used as frozen backbones and trained for the VP task.
For 2D, we evaluate the robot base placement method AIS~\cite{mees20icra_placements} and, following the NYC-Indoor-VPR dataset~\cite{Sheng_2024_NYC_Indoor_VPR}, the detection method AnyLoc-VLAD DINOv2~\cite{keetha2023anyloc}. We further benchmark two leading segmentation methods: SAM~\cite{kirillov2023segment} and SigLIP-2~\cite{tschannen2025siglip2multilingualvisionlanguage}.
In addition, we report results for CLIP-UNet~\cite{ramrakhya2024seeingtheunseen}, which, to the best of our knowledge, is the only existing VP method that can be directly applied to our benchmark. For fairness, CLIP-UNet was trained and evaluated on our dataset.
For 3D, we follow the detection methods used for evaluation in LiDAR-CS~\cite{fang2023lidarcs}, including PointPillars~\cite{Lang_2019_CVPR}, which transforms points into vertical pillars to form a 2D feature map; SECOND~\cite{yan2018second}, which introduces sparse 3D convolutions to replace traditional dense 3D convolutions; and PointRCNN~\cite{shi2019pointrcnn}, a point-based framework that uses PointNet++\cite{qi2017pointnet++} as its backbone to extract features for segmentation. We also include LPS-Net\cite{Liu_2024_ICRA_LPSNet}, a cloud-based place recognition system.
On top of each backbone (for both 2D and 3D), we train a placement head composed of a fully-connected layer to predict the binary VP maps. All reported results are based on the full benchmark (real + synthetic). Each result represents the average of three independent runs, with a standard deviation of less than $1\%$.

Table~\ref{tbl: full res} shows the VP results of the different methods on the Assistant Placement Aria benchmark.
In 2D, CLIP-UNet~\cite{ramrakhya2024seeingtheunseen} achieves the best results.
This is not surprising as virtual placement requires both segmentation and localization (detection) capabilities which CLIP-UNet combines while other methods  such as SAM~\cite{kirillov2023segment} and
AnyLoc-VLAD DINOv2~\cite{keetha2023anyloc} have but one.
In 3D, LPS-Net~\cite{Liu_2024_ICRA_LPSNet} attains the highest IoU, probably because it is a segmentation system while the other methods are detection systems. 

\begin{figure}[t]
\centering
\begin{tabular}{@{}c@{\hspace{2pt}}c@{\hspace{2pt}}c@{\hspace{2pt}}c@{\hspace{2pt}}c@{}}
 
  \rotatebox{90}{2D-Panel} & 
  \includegraphics[width=0.11\textwidth]{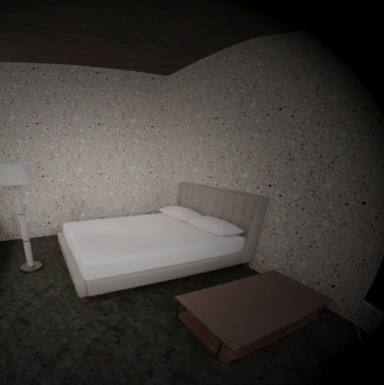} & 
  \includegraphics[width=0.11\textwidth]{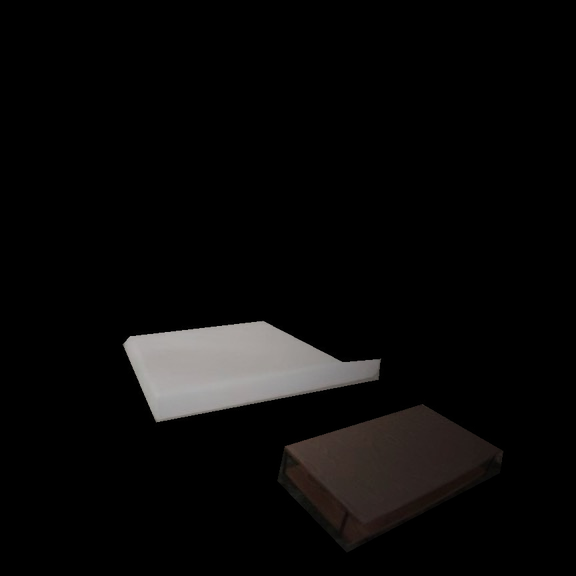} & 
  \includegraphics[width=0.11\textwidth]{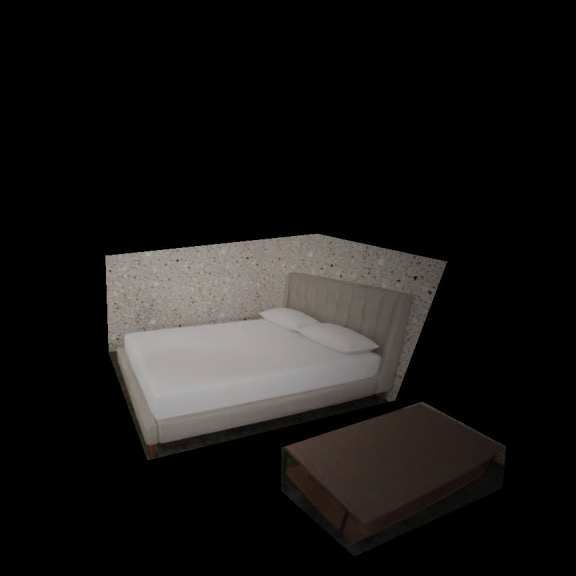} & 
  \includegraphics[width=0.11\textwidth]{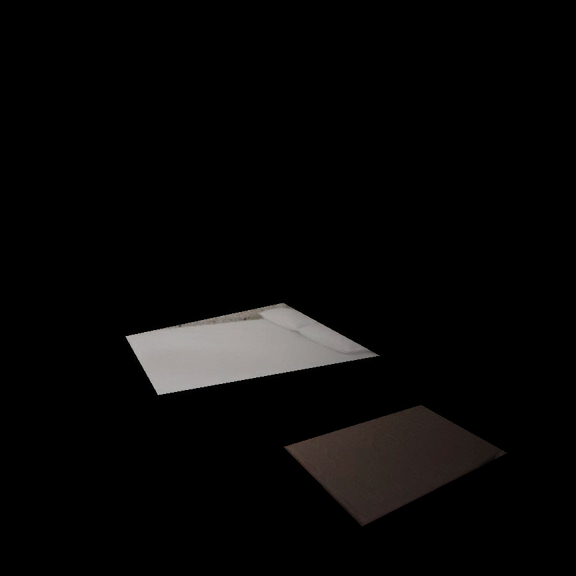}
  \\
  
 \rotatebox{90}{Sitting} & 
 \includegraphics[width=0.11\textwidth]{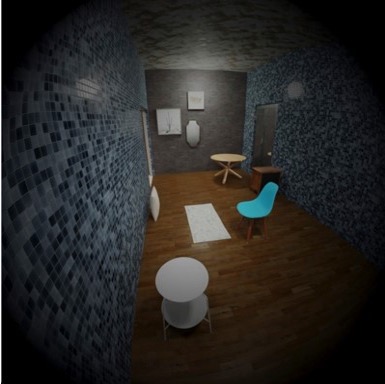} & 
 \includegraphics[width=0.11\textwidth]{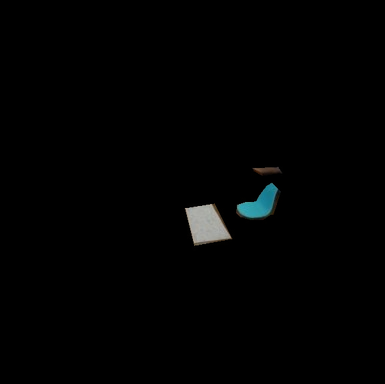} &
 \includegraphics[width=0.11\textwidth]{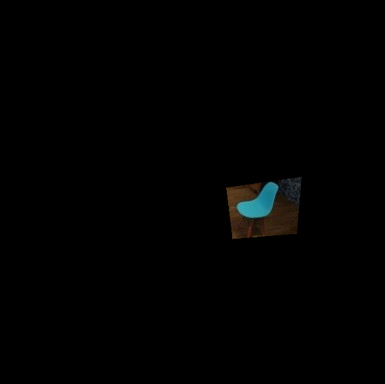} & 
 \includegraphics[width=0.11\textwidth]{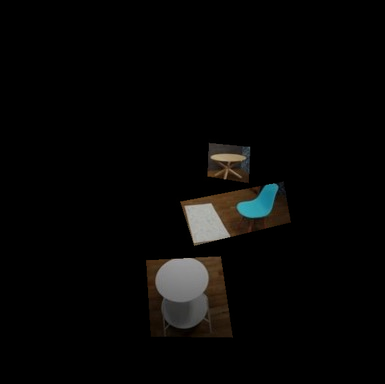} 
  \\
\rotatebox{90}{TV} & 
  \includegraphics[width=0.11\textwidth]{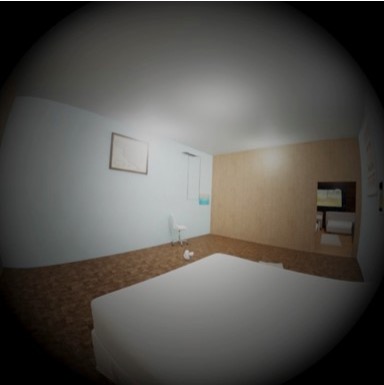} & 
  \includegraphics[width=0.11\textwidth]{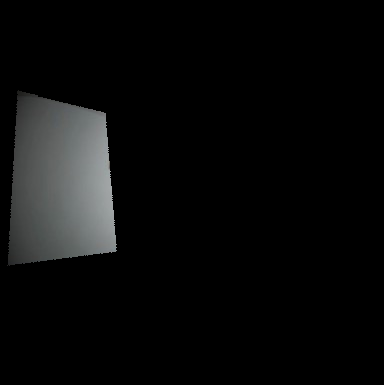} &
  \includegraphics[width=0.11\textwidth]{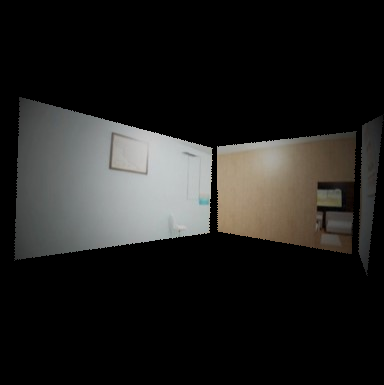} & 
  \includegraphics[width=0.11\textwidth]{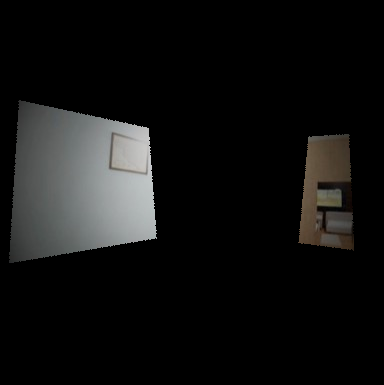} 
  \\
  
 & (a) RGB & (d) SAM & (f) AnyLoc & (g) CLIP-UNet \\
\end{tabular}
\caption{
{\bf Qualitative results.} 
Three examples of the placement performed by different 2D models. 
}
\label{fig:quantitative 2D different models}
\end{figure}

The performance on the full dataset is significantly higher than on the subset of manually tagged scenes across all methods. 
This is because the manually tagged scenes focus on the specific parts of each object that are placeable (e.g., the seat of a couch), rather than the entire object. 
As a result, detection and segmentation methods, which treat objects as wholes, perform significantly better on the automatically tagged scenes.
Fig.~\ref{fig:quantitative 2D different models} illustrates this effect by presenting qualitative VP results overlaid on the input images, showcasing predictions of both segmentation and detection methods. 
While AnyLoc-VLAD DINOv2 and SAM treat each object as a single entity, as evident in the 2D Panel task, where an entire table is predicted as a valid region rather than just the tabletop, CLIP-UNet identifies more specific placement areas.
This highlights the importance of fine-grained, preference-aware annotations for accurately evaluating VP methods.

\section{Conclusions}

In this paper, we addressed the challenge of virtual placement, a task that has remained underexplored due to the lack of suitable training data, and which has potential applications across various robotic fields, including assistive navigation and object placement.
We introduced annotations for the ASE~\cite{engel2023project} and AEO~\cite{straub2024efm3dbenchmarkmeasuringprogress} datasets, covering three distinct VP tasks: 2D Panel Placement, Sitting Suggestion, and TV Placement. Our annotation process includes both 2D and 3D labeling, along with per-object text descriptions.
To the best of our knowledge, this is the first benchmark to comprehensively address the diverse challenges and constraints inherent to virtual placement, including global, local, and human-centric considerations. We also evaluated several detection and segmentation models on our benchmark, establishing baselines for future research.
Future work may explore the development of new methods for both 2D and 3D virtual placement, leveraging the benchmark to advance human-centric scene understanding.

\section{Appendix}
\label{sec:appendix}

{\bf On the prompting process for the scene’s textual description.}
To generate the description of each object, we first identify a representative image patch that can serve as input for text generation. 
During preprocessing, we traverse the entire scene and record the frame in which each object instance occupies the largest image area. 
For that frame, we then extract a patch by cropping the instance segmentation with an additional $10$-px padding.
We found that this approach obtains acceptable object descriptions in the scene for our placement tasks. 
This description is then used for all other frames in which the object appear.

Furthermore, during the preprocessing step for ASE~\cite{engel2023project}, we created 3D bounding boxes for each object to enable extending the dataset to 3D scenarios. 
For each frame, we reprojected the ground-truth depth, pose, and camera intrinsics (6-DoF) to generate a point cloud. 
The instance segmentation was preserved for every reprojected point. 
To improve memory and computational efficiency, we resampled each frame’s point cloud using QuickFPS~\cite{han2023quickfps}, retaining only $500{,}000$ points. 
After processing all frames, we aligned the global point cloud with the $z$-axis using the floor segmentation. 
For each object instance, and disregarding the $z$-axis, we computed the $x$ and $y$ rotation and scale by finding the 2D bounding box that encloses the object’s points via their convex hull. 
Since the point cloud is aligned with the $z$-axis, the bounding box height was obtained directly as the difference between the highest and lowest points, without requiring rotation on that axis. 
Finally, we saved the 3D bounding box corresponding to the highest-area frame used for generating the text description of the 3D scenes, thereby enhancing the original dataset. 
These 3D bounding boxes can support a variety of applications beyond object placement.

 \bibliographystyle{IEEEtran}
 \bibliography{main}

@STRING{IEEE_J_CAD        = "{IEEE} Transactions on Computer-Aided Design of Integrated Circuits and Systems"}

@STRING{IEEE_J_RA         = "{IEEE} Transactions on Robotics and Automation"}

@string{cvpr="IEEE/CVF Conf. Computer Vision and Pattern Recognition (CVPR)"}

@string{cvpr_old="IEEE Conf. Computer Vision and Pattern Recognition (CVPR)"}

@string{iccv="IEEE/CVF Int'l Conf. Computer Vision (ICCV)"}

@string{iccv_old="IEEE Int'l Conf. Computer Vision (ICCV)"}

@string{eccv="European Conf. Computer Vision (ECCV)"}

@string{icra="IEEE Int'l Conf. Robotics and Automation (ICRA)"}

@string{iros="IEEE/RSJ Int'l Conf. Intelligent Robots and Systems (IROS)"}

@string{dv="Int'l Conf. 3D Vision (3DV)"}

@string{nips="Advances in Neural Information Processing Systems (NeurIPS)"}

@string{ram="IEEE Robotics Automation Magazine (RA-M)"}

@string{ijrr="The International Journal of Robotics Research (IJRR)"}

@STRING{IEEE_J_RA         = "{IEEE} Trans. Robot. Automat."}

@STRING{IEEE_J_CAD        = "{IEEE} Trans. Computer-Aided Design"}

@article{qi2017pointnet++,
  title={Pointnet++: Deep hierarchical feature learning on point sets in a metric space},
  author={Qi, Charles Ruizhongtai and Yi, Li and Su, Hao and Guibas, Leonidas J},
  journal=nips,
  volume={30},
  year={2017}
}

@article{engel2023project,
  title={Project aria: A new tool for egocentric multi-modal ai research},
  author={Engel, Jakob and Somasundaram, Kiran and Goesele, Michael and Sun, Albert and Gamino, Alexander and Turner, Andrew and Talattof, Arjang and Yuan, Arnie and Souti, Bilal and Meredith, Brighid and others},
  journal={arXiv preprint arXiv:2308.13561},
  year={2023}
}

@inproceedings{pan2023aria,
  title={Aria digital twin: A new benchmark dataset for egocentric 3d machine perception},
  author={Pan, Xiaqing and Charron, Nicholas and Yang, Yongqian and Peters, Scott and Whelan, Thomas and Kong, Chen and Parkhi, Omkar and Newcombe, Richard and Ren, Yuheng Carl},
  booktitle=iccv,
  pages={20133--20143},
  year={2023}
}

@inproceedings{kirillov2023segment,
  title={Segment anything},
  author={Kirillov, Alexander and Mintun, Eric and Ravi, Nikhila and Mao, Hanzi and Rolland, Chloe and Gustafson, Laura and Xiao, Tete and Whitehead, Spencer and Berg, Alexander C and Lo, Wan-Yen and others},
  booktitle=iccv,
  pages={4015--4026},
  year={2023}
}

@article{replica19arxiv,
  title =   {The {R}eplica Dataset: A Digital Replica of Indoor Spaces},
  author =  {Julian Straub and Thomas Whelan and Lingni Ma and Yufan Chen and Erik Wijmans and Simon Green and Jakob J. Engel and Raul Mur-Artal and Carl Ren and Shobhit Verma and Anton Clarkson and Mingfei Yan and Brian Budge and Yajie Yan and Xiaqing Pan and June Yon and Yuyang Zou and Kimberly Leon and Nigel Carter and Jesus Briales and  Tyler Gillingham and  Elias Mueggler and Luis Pesqueira and Manolis Savva and Dhruv Batra and Hauke M. Strasdat and Renzo De Nardi and Michael Goesele and Steven Lovegrove and Richard Newcombe },
  journal = {arXiv preprint arXiv:1906.05797},
  year =    {2019}
}

@article{Matterport3D,
  title={Matterport3D: Learning from RGB-D Data in Indoor Environments},
  author={Chang, Angel and Dai, Angela and Funkhouser, Thomas and Halber, Maciej and Niessner, Matthias and Savva, Manolis and Song, Shuran and Zeng, Andy and Zhang, Yinda},
  journal=dv,
  year={2017}
}

@inproceedings{dai2017scannet,
    title={ScanNet: Richly-annotated 3D Reconstructions of Indoor Scenes},
    author={Dai, Angela and Chang, Angel X. and Savva, Manolis and Halber, Maciej and Funkhouser, Thomas and Nie{\ss}ner, Matthias},
    booktitle = cvpr,
    year = {2017}
}

@inproceedings{song2015sun,
  title={Sun rgb-d: A rgb-d scene understanding benchmark suite},
  author={Song, Shuran and Lichtenberg, Samuel P and Xiao, Jianxiong},
  booktitle=cvpr_old,
  pages={567--576},
  year={2015}
}

@inproceedings{roberts2021hypersim,
  title={Hypersim: A photorealistic synthetic dataset for holistic indoor scene understanding},
  author={Roberts, Mike and Ramapuram, Jason and Ranjan, Anurag and Kumar, Atulit and Bautista, Miguel Angel and Paczan, Nathan and Webb, Russ and Susskind, Joshua M},
  booktitle=iccv,
  pages={10912--10922},
  year={2021}
}

@inproceedings{li2021openrooms,
  title={Openrooms: An open framework for photorealistic indoor scene datasets},
  author={Li, Zhengqin and Yu, Ting-Wei and Sang, Shen and Wang, Sarah and Song, Meng and Liu, Yuhan and Yeh, Yu-Ying and Zhu, Rui and Gundavarapu, Nitesh and Shi, Jia and others},
  booktitle=cvpr,
  pages={7190--7199},
  year={2021}
}

@article{chen2020scanrefer,
    title={ScanRefer: 3D Object Localization in RGB-D Scans using Natural Language},
    author={Chen, Dave Zhenyu and Chang, Angel X and Nie{\ss}ner, Matthias},
    journal={16th European Conference on Computer Vision (ECCV)},
    year={2020}
}

@InProceedings{Peng_2023_CVPR,
    author    = {Peng, Songyou and Genova, Kyle and Jiang, Chiyu {\textquotedblleft}Max{\textquotedblright} and Tagliasacchi, Andrea and Pollefeys, Marc and Funkhouser, Thomas},
    title     = {OpenScene: 3D Scene Understanding With Open Vocabularies},
    booktitle = cvpr,
    month     = {June},
    year      = {2023},
    pages     = {815-824}
}

@InProceedings{Ding_2023_CVPR,
    author    = {Ding, Runyu and Yang, Jihan and Xue, Chuhui and Zhang, Wenqing and Bai, Song and Qi, Xiaojuan},
    title     = {PLA: Language-Driven Open-Vocabulary 3D Scene Understanding},
    booktitle = cvpr,
    month     = {June},
    year      = {2023},
    pages     = {7010-7019}
}

@InProceedings{Azuma_2022_CVPR,
    author    = {Azuma, Daichi and Miyanishi, Taiki and Kurita, Shuhei and Kawanabe, Motoaki},
    title     = {ScanQA: 3D Question Answering for Spatial Scene Understanding},
    booktitle = cvpr,
    month     = {June},
    year      = {2022},
    pages     = {19129-19139}
}

@misc{straub2024efm3dbenchmarkmeasuringprogress,
      title={EFM3D: A Benchmark for Measuring Progress Towards 3D Egocentric Foundation Models}, 
      author={Julian Straub and Daniel DeTone and Tianwei Shen and Nan Yang and Chris Sweeney and Richard Newcombe},
      year={2024},
}

@InProceedings{Zhu_2023_CVPR,
    author    = {Zhu, Sijie and Lin, Zhe and Cohen, Scott and Kuen, Jason and Zhang, Zhifei and Chen, Chen},
    title     = {TopNet: Transformer-Based Object Placement Network for Image Compositing},
    booktitle = cvpr,
    month     = {June},
    year      = {2023},
    pages     = {1838-1847}
}

@article{DBLP:journals/corr/abs-2107-01889,
  author       = {Liu Liu and
                  Bo Zhang and
                  Jiangtong Li and
                  Li Niu and
                  Qingyang Liu and
                  Liqing Zhang},
  title        = {{OPA:} Object Placement Assessment Dataset},
  journal      = {CoRR},
  volume       = {abs/2107.01889},
  year         = {2021},
  eprinttype    = {arXiv},
}

@article{niu2021making,
  title={Making images real again: A comprehensive survey on deep image composition},
  author={Niu, Li and Cong, Wenyan and Liu, Liu and Hong, Yan and Zhang, Bo and Liang, Jing and Zhang, Liqing},
  journal={arXiv preprint arXiv:2106.14490},
  year={2021}
}

@inproceedings{lu2023tf,
  title={Tf-icon: Diffusion-based training-free cross-domain image composition},
  author={Lu, Shilin and Liu, Yanzhu and Kong, Adams Wai-Kin},
  booktitle=iccv,
  pages={2294--2305},
  year={2023}
}

@inproceedings{zhou2022learning,
  title={Learning object placement via dual-path graph completion},
  author={Zhou, Siyuan and Liu, Liu and Niu, Li and Zhang, Liqing},
  booktitle=eccv,
  pages={373--389},
  year={2022},
  organization={Springer}
}

@misc{meng2023interactiveobject,
    author = {Quanling Meng and Qinglin Liu},
    title = {Interactive Object Placement with Reinforcement Learning},
    year = 2023,
}

@misc{ramrakhya2024seeingtheunseen,
    author = {Ram Ramrakhya and Aniruddha Kembhavi and Dhruv Batra and Zsolt Kira and Kuo-Hao Zeng and Luca Weihs},
    title = {Seeing the Unseen: Visual Common Sense for Semantic Placement},
    year = 2024,
}

@misc{lee2018contextawaresynthesis,
    author = {Donghoon Lee and Sifei Liu and Jinwei Gu and Ming-Yu Liu and Ming-Hsuan Yang and Jan Kautz},
    title = {Context-Aware Synthesis and Placement of Object Instances},
    year = 2018,
}

@article{schuhmann2022laion,
  title={Laion-5b: An open large-scale dataset for training next generation image-text models},
  author={Schuhmann, Christoph and Beaumont, Romain and Vencu, Richard and Gordon, Cade and Wightman, Ross and Cherti, Mehdi and Coombes, Theo and Katta, Aarush and Mullis, Clayton and Wortsman, Mitchell and others},
  journal=nips,
  volume={35},
  pages={25278--25294},
  year={2022}
}

@inproceedings{khanna2024habitat,
  title={Habitat synthetic scenes dataset (hssd-200): An analysis of 3d scene scale and realism tradeoffs for objectgoal navigation},
  author={Khanna, Mukul and Mao, Yongsen and Jiang, Hanxiao and Haresh, Sanjay and Shacklett, Brennan and Batra, Dhruv and Clegg, Alexander and Undersander, Eric and Chang, Angel X and Savva, Manolis},
  booktitle=cvpr,
  pages={16384--16393},
  year={2024}
}

@inproceedings{liu2023llava,
    author      = {Liu, Haotian and Li, Chunyuan and Wu, Qingyang and Lee, Yong Jae},
    title       = {Visual Instruction Tuning},
    booktitle   = nips,
    year        = {2023}
}

@inproceedings{parihar2024text2place,
  title={Text2Place: Affordance-Aware Text Guided Human Placement},
  author={Parihar, Rishubh and Gupta, Harsh and VS, Sachidanand and Babu, R Venkatesh},
  booktitle=eccv,
  pages={57--77},
  year={2024},
  organization={Springer}
}

@article{han2023quickfps,
  title={QuickFPS: Architecture and Algorithm Co-Design for Farthest Point Sampling in Large-Scale Point Clouds},
  author={Han, Meng and Wang, Liang and Xiao, Limin and Zhang, Hao and Zhang, Chenhao and Xu, Xiangrong and Zhu, Jianfeng},
  journal=IEEE_J_CAD,
  year={2023},
  publisher={IEEE}
}

@misc{tschannen2025siglip2multilingualvisionlanguage,
      title={SigLIP 2: Multilingual Vision-Language Encoders with Improved Semantic Understanding, Localization, and Dense Features}, 
      author={Michael Tschannen and Alexey Gritsenko and Xiao Wang and Muhammad Ferjad Naeem and Ibrahim Alabdulmohsin and Nikhil Parthasarathy and Talfan Evans and Lucas Beyer and Ye Xia and Basil Mustafa and Olivier Hénaff and Jeremiah Harmsen and Andreas Steiner and Xiaohua Zhai},
      year={2025},
}

@misc{yoffe2023octopusopenvocabularycontenttracking,
      title={OCTOPUS: Open-vocabulary Content Tracking and Object Placement Using Semantic Understanding in Mixed Reality}, 
      author={Luke Yoffe and Aditya Sharma and Tobias Höllerer},
      year={2023},
}

@inproceedings{10.1145/3551349.3561160,
author = {Rafi, Tahmid and Zhang, Xueling and Wang, Xiaoyin},
title = {PredART: Towards Automatic Oracle Prediction of Object Placements in Augmented Reality Testing},
year = {2023},
isbn = {9781450394758},
publisher = {Association for Computing Machinery},
address = {New York, NY, USA},
booktitle = {Proceedings of the 37th IEEE/ACM International Conference on Automated Software Engineering},
articleno = {77},
numpages = {13},
location = {Rochester, MI, USA},
series = {ASE '22}
}

@INPROCEEDINGS{9811600,
  author={Birr, Timo and Pohl, Christoph and Asfour, Tamim},
  booktitle=icra, 
  title={Oriented Surface Reachability Maps for Robot Placement}, 
  year={2022},
  volume={},
  number={},
  pages={3357-3363}}

@INPROCEEDINGS{10801373,
  author={Wachter, Alexander and Kugi, Andreas and Hartl-Nesic, Christian},
  booktitle=iros, 
  title={Time-Optimal TCP and Robot Base Placement for Pick-and-Place Tasks in Highly Constrained Environments}, 
  year={2024},
  volume={},
  number={},
  pages={2251-2257}}

@INPROCEEDINGS{9560853,
  author={Yang, Tong and Miro, Jaime Valls and Wang, Yue and Xiong, Rong},
  booktitle=icra, 
  title={Optimal Object Placement for Minimum Discontinuity Non-revisiting Coverage Task}, 
  year={2021},
  volume={},
  number={},
  pages={8422-8428}}

@inproceedings{mees20icra_placements,
author = {Oier Mees and Alp Emek and Johan Vertens and Wolfram Burgard},
title = {Learning Object Placements For Relational Instructions by Hallucinating Scene Representations},
booktitle = icra,
year = 2020,
address = {Paris, France}
}

@article{keetha2023anyloc,
  title={AnyLoc: Towards Universal Visual Place Recognition}, 
  author={Keetha, Nikhil and Mishra, Avneesh and Karhade, Jay and Jatavallabhula, Krishna Murthy and Scherer, Sebastian and Krishna, Madhava and Garg, Sourav},
  journal=IEEE_J_RA,
  year={2023},
  publisher={IEEE},
  volume={9},
  number={2},
  pages={1286-1293},
}

@inproceedings{fang2023lidarcs,
    title={LiDAR-CS Dataset: LiDAR Point Cloud Dataset with Cross-Sensors for 3D Object Detection},
    author={Jin Fang and Dingfu Zhou and Jingjing Zhao and Chenming Wu and Chulin Tang and Cheng-Zhong Xu and Liangjun Zhang},
    booktitle=icra,
    year={2024}
}

@InProceedings{Lang_2019_CVPR,
  author    = {Lang, Alex H. and Vora, Sourabh and Caesar, Holger and Zhou, Lubing and Yang, Jiong and Beijbom, Oscar},
  title     = {PointPillars: Fast Encoders for Object Detection From Point Clouds},
  booktitle = cvpr,
  year      = {2019},
  month     = {June},
  pages     = {12697--12705},
  publisher = {Computer Vision Foundation / IEEE}
}

@article{yan2018second,
  title={SECOND: Sparsely Embedded Convolutional Detection},
  author={Yan, Yan and Mao, Yuxing and Li, Bo},
  journal={Sensors},
  volume={18},
  number={10},
  pages={3337},
  year={2018},
  publisher={MDPI},
}

@InProceedings{shi2019pointrcnn,
  title     = {PointRCNN: 3D Object Proposal Generation and Detection from Point Cloud},
  author    = {Shi, Shaoshuai and Wang, Xiaogang and Li, Hongsheng},
  booktitle = cvpr,
  year      = {2019},
  pages     = {770--779}
}

@InProceedings{Liu_2024_ICRA_LPSNet,
  author    = {Liu, Chengxin and Chen, Guiyou and Song, Ran},
  title     = {LPS‐Net: Lightweight Parameter‐Shared Network for Point Cloud‐Based Place Recognition},
  booktitle = icra,
  year      = {2024},
  pages     = {448--454},
}

@InProceedings{Sheng_2024_NYC_Indoor_VPR,
  author    = {Sheng, Diwei and Yang, Anbang and Rizzo, John-Ross and Feng, Chen},
  title     = {NYC-Indoor-VPR: A Long-Term Indoor Visual Place Recognition Dataset with Semi-Automatic Annotation},
  booktitle = icra,
  year      = {2024},
  pages     = {14853--14859},
}

@article{jiang2012learning_place,
  title={Learning to place new objects in a scene},
  author={Jiang, Yun and Lim, Marcus and Zheng, Changxi and Saxena, Ashutosh},
  journal=ijrr,
  volume={31},
  number={9},
  pages={1021--1043},
  year={2012},
  publisher={SAGE Publications Sage UK: London, England}
}

@article{jiang2012learning_arrangements,
  title={Learning object arrangements in 3d scenes using human context},
  author={Jiang, Yun and Lim, Marcus and Saxena, Ashutosh},
  journal={arXiv preprint arXiv:1206.6462},
  year={2012}
}

@inproceedings{jund2018optimization,
  title={Optimization beyond the convolution: Generalizing spatial relations with end-to-end metric learning},
  author={Jund, Philipp and Eitel, Andreas and Abdo, Nichola and Burgard, Wolfram},
  booktitle=icra,
  pages={1--7},
  year={2018},
  organization={IEEE}
}

@inproceedings{zampogiannis2015learning,
  title={Learning the spatial semantics of manipulation actions through preposition grounding},
  author={Zampogiannis, Konstantinos and Yang, Yezhou and Fermuller, Cornelia and Aloimonos, Yiannis},
  booktitle=icra,
  pages={1389--1396},
  year={2015},
  organization={IEEE}
}

@inproceedings{dwibedi2017cut,
  title={Cut, paste and learn: Surprisingly easy synthesis for instance detection},
  author={Dwibedi, Debidatta and Misra, Ishan and Hebert, Martial},
  booktitle=iccv_old,
  pages={1301--1310},
  year={2017}
}

@inproceedings{lee2018context,
  title={Context-aware synthesis and placement of object instances},
  author={Lee, Donghoon and Liu, Sifei and Gu, Jinwei and Liu, Ming-Yu and Yang, Ming-Hsuan and Kautz, Jan},
  booktitle=nips,
  pages={10393--10403},
  year={2018}
}

@ARTICLE{7339473,
  author={Lowry, Stephanie and Sünderhauf, Niko and Newman, Paul and Leonard, John J. and Cox, David and Corke, Peter and Milford, Michael J.},
  journal=IEEE_J_RA, 
  title={Visual Place Recognition: A Survey}, 
  year={2016},
  volume={32},
  number={1},
  pages={1-19}}

\end{document}